\documentclass[sigconf]{acmart}

\usepackage{algorithmic}
\usepackage[linesnumbered, ruled, vlined]{algorithm2e}
\def\header{\vspace{0.8mm} \noindent}
\usepackage{float}
\usepackage{enumerate}
\usepackage{enumitem}

\AtBeginDocument{%
  \providecommand\BibTeX{{%
    \normalfont B\kern-0.5em{\scshape i\kern-0.25em b}\kern-0.8em\TeX}}}

\setcopyright{acmlicensed}

\copyrightyear{2025}
\acmYear{2025}
\setcopyright{acmlicensed}
\acmConference[KDD '25] {Proceedings of the 31st ACM SIGKDD Conference on Knowledge Discovery and Data Mining V.1}{August 3--7, 2025}{Toronto, ON, Canada.}
\acmBooktitle{Proceedings of the 31st ACM SIGKDD Conference on Knowledge Discovery and Data Mining V.1 (KDD '25), August 3--7, 2025, Toronto, ON, Canada}
\acmISBN{979-8-4007-1245-6/25/08}
\acmDOI{10.1145/XXXXXX.XXXXXX}

\begin{document}

%%
%% The "title" command has an optional parameter,
%% allowing the author to define a "short title" to be used in page headers.
\title{Large-Scale Spectral Graph Neural Networks via Laplacian Sparsification: Technical Report}

%%
%% The "author" command and its associated commands are used to define
%% the authors and their affiliations.
%% Of note is the shared affiliation of the first two authors, and the
%% "authornote" and "authornotemark" commands
%% used to denote shared contribution to the research.
% \author{Ben Trovato}
% \authornote{Both authors contributed equally to this research.}
% \email{trovato@corporation.com}
% \orcid{1234-5678-9012}
% \author{G.K.M. Tobin}
% \authornotemark[1]
% \email{webmaster@marysville-ohio.com}
% \affiliation{%
%   \institution{Institute for Clarity in Documentation}
%   \streetaddress{P.O. Box 1212}
%   \city{Dublin}
%   \state{Ohio}
%   \country{USA}
%   \postcode{43017-6221}
% }

\author{Haipeng Ding}
\email{dinghaipeng@ruc.edu.cn}
\affiliation{%
    \institution{Renmin University of China}
    \city{Beijing}
    \country{China}
}

\author{Zhewei Wei}
\authornote{Zhewei Wei is the corresponding author.}
\email{zhewei@ruc.edu.cn}
\affiliation{%
    \institution{Renmin University of China}
    \city{Beijing}
    \country{China}
}

\author{Yuhang Ye}
\email{yeyuhang@huawei.com}
\affiliation{%
    \institution{Huawei Poisson Lab}
    \city{Shenzhen}
    \state{Guangdong}
    \country{China}
}

% \author{Lars Th{\o}rv{\"a}ld}
% \affiliation{%
%   \institution{The Th{\o}rv{\"a}ld Group}
%   \streetaddress{1 Th{\o}rv{\"a}ld Circle}
%   \city{Hekla}
%   \country{Iceland}}
% \email{larst@affiliation.org}

% \author{Valerie B\'eranger}
% \affiliation{%
%   \institution{Inria Paris-Rocquencourt}
%   \city{Rocquencourt}
%   \country{France}
% }

% \author{Aparna Patel}
% \affiliation{%
%  \institution{Rajiv Gandhi University}
%  \streetaddress{Rono-Hills}
%  \city{Doimukh}
%  \state{Arunachal Pradesh}
%  \country{India}}

% \author{Huifen Chan}
% \affiliation{%
%   \institution{Tsinghua University}
%   \streetaddress{30 Shuangqing Rd}
%   \city{Haidian Qu}
%   \state{Beijing Shi}
%   \country{China}}

% \author{Charles Palmer}
% \affiliation{%
%   \institution{Palmer Research Laboratories}
%   \streetaddress{8600 Datapoint Drive}
%   \city{San Antonio}
%   \state{Texas}
%   \country{USA}
%   \postcode{78229}}
% \email{cpalmer@prl.com}

% \author{John Smith}
% \affiliation{%
%   \institution{The Th{\o}rv{\"a}ld Group}
%   \streetaddress{1 Th{\o}rv{\"a}ld Circle}
%   \city{Hekla}
%   \country{Iceland}}
% \email{jsmith@affiliation.org}

% \author{Julius P. Kumquat}
% \affiliation{%
%   \institution{The Kumquat Consortium}
%   \city{New York}
%   \country{USA}}
% \email{jpkumquat@consortium.net}

%%
%% By default, the full list of authors will be used in the page
%% headers. Often, this list is too long, and will overlap
%% other information printed in the page headers. This command allows
%% the author to define a more concise list
%% of authors' names for this purpose.
\renewcommand{\shortauthors}{Trovato and Tobin, et al.}

%%
%% The abstract is a short summary of the work to be presented in the
%% article.
\begin{abstract}
Graph Neural Networks (GNNs) play a pivotal role in graph-based tasks for their proficiency in representation learning.
Among the various GNN methods, spectral GNNs employing polynomial filters have shown promising performance on tasks involving both homophilous and heterophilous graph structures.
However, The scalability of spectral GNNs on large graphs is limited because they learn the polynomial coefficients through multiple forward propagation executions during forward propagation.
Existing works have attempted to scale up spectral GNNs by eliminating the linear layers on the input node features, a change that can disrupt end-to-end training, potentially impact performance, and become impractical with high-dimensional input features.
%Current scalable spectral methods detach the graph propagation and linear layers, and preprocess the message-passing phase.
%While this operation enhances scalability, it introduces a disconnection in end-to-end training, potentially affecting performance and incurring additional memory costs.
% On the other hand, scalable spectral GNNs detach the graph propagation and linear layers, allowing the message-passing phase to be pre-computed and ensuring effective scalability on large graphs. 
% However, this pre-computation can disrupt end-to-end training, possibly impacting performance, and becomes impractical when dealing with high-dimensional input features.
%1) Existing techniques designed to enhance GNN scalability face limitations due to the multi-step message-passing of spectral GNNs.
%2) Detaching the graph propagation and the linear layers may make the training scalable but can compromise the performance of the GNN models and disrupt end-to-end training.
To address the above challenges, we propose ``Spectral Graph Neural Networks with Laplacian Sparsification (SGNN-LS)'', a novel graph spectral sparsification method to approximate the propagation patterns of spectral GNNs.
We prove that our proposed method generates Laplacian sparsifiers that can approximate both fixed and learnable polynomial filters with theoretical guarantees.
%By considering multi-hop neighbor interactions into one-hop operations, our approach facilitates the use of scalable techniques.
Our method allows the application of linear layers on the input node features, enabling end-to-end training as well as the handling of raw text features.
We conduct an extensive experimental analysis on datasets spanning various graph scales and properties to demonstrate the superior efficiency and effectiveness of our method.
The results show that our method yields superior results in comparison with the corresponding approximated base models, especially on dataset Ogbn-papers100M(111M nodes, 1.6B edges) and MAG-scholar-C (2.8M features).
% Impressively, our approach also performs well on large-scale datasets like Ogbn-papers100M with $10^8$ nodes and $10^9$ edges.
\end{abstract}

%%
%% The code below is generated by the tool at http://dl.acm.org/ccs.cfm.
%% Please copy and paste the code instead of the example below.
%%


%%
%% Keywords. The author(s) should pick words that accurately describe
%% the work being presented. Separate the keywords with commas.

%% A "teaser" image appears between the author and affiliation
%% information and the body of the document, and typically spans the
%% page.
% \begin{teaserfigure}
%   \includegraphics[width=\textwidth]{sampleteaser}
%   \caption{Seattle Mariners at Spring Training, 2010.}
%   \Description{Enjoying the baseball game from the third-base
%   seats. Ichiro Suzuki preparing to bat.}
%   \label{fig:teaser}
% \end{teaserfigure}

% \received{20 February 2007}
% \received[revised]{12 March 2009}
% \received[accepted]{5 June 2009}

%%
%% This command processes the author and affiliation and title
%% information and builds the first part of the formatted document.
\maketitle

\section{Introduction}
\label{sec:intro}
%Graphs serve as a widely-used data structure in real-world scenarios, owing to their efficient and intuitive modeling of complex relationships.
Graph Neural Networks (GNNs) have gathered increasing research attention because of their versatility in handling graph-structured data.
They have demonstrated prominent performance in several kinds of real-world graph learning tasks, including link prediction~\citep{DBLP:conf/www/SankarLYS21}, recommendation systems~\citep{DBLP:recommendationsystem1, DBLP:recommendationsystem2, DBLP:recommendationsystem3}, social analysis~\citep{DBLP:socialanalysis1}, drug discovery~\citep{DBLP:drugdiscovery1}, and traffic forecasting~\citep{DBLP:trafficforcasting1}.
GNNs can be broadly categorized into spatial GNNs, including GCN~\citep{DBLP:GCN}, GAT~\citep{DBLP:GAT}, GCNII~\citep{DBLP:GCNII}, GIN~\citep{DBLP:GIN}, MGNN~\citep{DBLP:MGNN}, and spectral GNNs.

Spectral GNNs represent one fundamental branch of GNNs that works by constructing a graph filter in the spectral domain of the graph Laplacian matrix.
This filtering mechanism enables the recombination of graph signals at different frequencies, effectively leveraging their spectral properties.
%Direct graph filter operation requires the eigendecomposition of the graph Laplacian matrix, whose time and space complexity are $O(n^3)$ and $O(n^2)$, respectively.
%Constrained by the impractical overhead of eigendecomposition, recent efforts have typically resorted to either predetermined graph convolution or polynomial bases with learnable coefficients to approximate the desired graph filtering operation.
Constrained by the impractical overhead of eigendecomposition, various works adopt distinct polynomial bases to approximate the desired filter operation, such as GPR-GNN~\citep{DBLP:GPR-GNN} leverages a monomial basis, BernNet~\citep{DBLP:BernNet} employs a Bernstein polynomial basis, JacobiConv~\citep{DBLP:JacobiConv} uses the Jacobi basis, and OptBasisGNN\citep{DBLP:OptBasis} learns optimal bases from the input signals.

In general, spectral GNNs employing polynomial filters can be formally expressed as $\mathbf{Y}=g_\mathbf{w}(\mathbf{L}, f_\theta(\mathbf{X}))$, where $g_\mathbf{w}(\cdot)$ denotes the polynomial graph filter with coefficients $\mathbf{w}$, $f_\theta(\cdot)$ represents the linear layer with learnable parameters $\theta$, $\mathbf{L}$ is the Laplacian matrix, $\mathbf{X}$ and $\mathbf{Y}$ refer to the original node representation matrix and the output, respectively. 
Unlike GNNs that design a uniform aggregation function, spectral GNNs use polynomial filters to combine representations from $K$-hop neighbors, where $K$ is the polynomial degree.
%Differing from most classic GNNs, which typically rely on neighborhood aggregation, spectral GNNs employing polynomial filters necessitate the representations from $K$-hop neighbor, where $K$ is the polynomial degree.
This property enables spectral GNNs to capture a broader scope of graphs and alleviates the dependence on ``homophily assumption''.

\header{\bf Motivation.} 
%Spectral GNNs have shown their superiority in tackling graph learning tasks, especially on heterophilous graphs.
Recent advancements in scalable GNN architectures, particularly those employing subsampling-based methods, have been primarily designed to promote the scalability of spatial GNNs, especially the vanilla GCN~\citep{DBLP:GCN}.
These scalable methods are not compatible with spectral GNNs due to the requirement of computing $\mathbf{L}^kf_\theta(\mathbf{X})$, which demands $k$ iterations of full graph propagation during training's forward propagation phase.
Moreover, existing acceleration devices, such as GPUs, suffer bottlenecks in storing the computational trees and node representation for extensive graphs.
Drawing inspirations from SGC~\citep{DBLP:SGC}, several spectral works such as ChebNetII~\citep{DBLP:ChebNetII} and OptBasis~\citep{DBLP:OptBasis} apply a trick to detach the graph propagation phase from the linear layers.
This trick enables the precomputation of $\mathbf{L}^k\mathbf{X}$, and converts the model training into the linear combination of $\mathbf{L}^k\mathbf{X}$ using learnable coefficient $w_k$ and linear layers.
However, this trick has lots of potential defects, including disrupting end-to-end training, negatively affecting performance, and becoming impractical when facing high-dimensional features.
% This process is independent of the graph structure and can be mini-batched naturally.
% Nevertheless, this method introduces the following defects with scalability: 
% 1) Some researches~\citep{DBLP:SGCxGCN1, DBLP:SGCxGCN2} argue that separating training from the graph structure simplifies the network architecture and will negatively impact performance. 
% 2) The training is no longer end-to-end due to the graph propagation preprocess.
% Emerging approaches~\citep{DBLP:LM1, DBLP:LM2} based on language models~\citep{DBLP:BERT, DBLP:GPT3} enhance the performance on Ogbn-papers100M, which requires the raw text as input and end-to-end training.
% 3) Preprocess is potentially impractical when dealing with large graphs with high-dimensional node features since the dimension of original features cannot be reduced by MLP.

Given these considerations, it prompts the question: \emph{Is there an approach to enhance the \textbf{scalability} of spectral GNNs \textbf{without decoupling} the graph propagation phase?}

\header{\bf Contribution.} 
We propose a novel methodology termed \textit{Spectral Graph Neural Networks with Laplacian Sparsification} (SGNN-LS), inspired by the classic technique of Laplacian sparsification.
This technique offers a strategy for deriving $\varepsilon$-sparsifiers of the equivalent propagation matrices associated with spectral GNNs with a high probability.
% Inspired by Laplacian sparsification, we propose a novel approach to approximate the equivalent propagation matrix of filters in spectral methods.
Specifically, our method approximates $\widetilde{\mathbf{L}}_K\approx\sum_{k=0}^Kw_k\mathbf{L}^k$, while ensuring the number of non-zeros in $\widetilde{\mathbf{L}}_K$ remains within $O\left(\frac{n\log n}{\varepsilon^2}\right)$.
Such sparsification not only facilitates the compression of multi-step graph propagation but also effectively connects multi-hop neighbors.
This, in turn, enables the application of various scalable GNN algorithms~\citep{DBLP:GraphSAGE, DBLP:LADIES, DBLP:GNNAutoScale}.
% Such sparsification effectively connects multi-hop neighbors by compressing the multi-step graph propagation, which enables the application of numerous scalable GNN algorithms~\citep{DBLP:GraphSAGE, DBLP:LADIES, DBLP:GNNAutoScale}.
% squeezes the multi-step graph propagation, effectively connecting multi-hop neighbors, and makes many scalable techniques of GNNs applicable.
Importantly, our approach retains the integration of graph propagation within the model.
Our contributions are summarized as follows:
\begin{itemize}[leftmargin = *]
    \item {\bf Plug-and-play strategy for scaling up spectral GNNs.} 
    We propose a method that is the first work tackling the scalability issue of \textbf{spectral GNNs} with either static or learnable polynomial coefficients to the best of our knowledge.
    % designed to approximate the equivalent propagation matrix of Laplacian filters. 
    This strategy offers adaptions of Laplacian sparsification suitable for different scenarios, including models with static polynomial coefficients, those with learnable polynomial coefficients, and a node-wise sampling way for semi-supervised tasks. Our codes are released at \url{https://anonymous.4open.science/r/SGNN-LS-release-B926/}.
    \item {\bf Theoretical analysis.}
    We present rigorous mathematical proofs to demonstrate that our methods construct Laplacian sparsifiers of dense matrix $\sum_{k=0}^K w_k\mathbf{L}^k$ for both static and learnable $w_k$ within $O\left(\frac{n\log n}{\varepsilon^2}\right)$ edges, a high probability $1-K/n$, and an approximation error $\varepsilon$.
    Additionally, we introduce a new loss function and establish that the relative error in the calculated loss between the propagated signals, using accurate and approximated propagation matrices, remains within $O(\varepsilon)$.
    These properties ensure the quality and reliability of generated sparsifiers, guaranteeing the robust performance and efficiency of our model.
    \item {\bf Extensive experiments.}
    We conduct comprehensive experiments to validate the effectiveness and scalability of our methods when applied to various spectral GNNs.
    The results consistently highlight the practical performance of our method, showcasing stable improvements compared to the corresponding baselines.
    Notably, our method enables the efficient training of GPR-GNN and APPNP on dataset Ogbn-papers100M (0.11B nodes, 1.62B edges) and MAG-Scholar-C (2.78M features), achieving commendable performance metrics.
    
\end{itemize}

\section{Preliminaries}
\label{sec:prelim}
\header{\bf Notations.}
In this study, we consider the undirected graph $G=(V, E)$, where $V$ represents the node set and $E$ is the edge set. 
Let $n=|V|$ and $m=|E|$ denote the number of nodes and edges, respectively.
$\mathbf{A}\in\{0,1\}^{n\times n}$ represents the adjacency matrix of the graph $G$. The diagonal degree matrix is denoted by $\mathbf{D}$, and $\mathbf{D}_{ii}=\sum_{j} \mathbf{A}_{ij}$.
%In certain scenarios, we augment the graph by adding the self-loops to all the nodes in $G$, leading to the self-looped graph $\widetilde{G}$, whose adjacency matrix $\widetilde{\mathbf{A}}=\mathbf{A}+\mathbf{I}_n$.
The normalized adjacency matrix and normalized graph Laplacian of $G$ are defined as $\mathbf{P}=\mathbf{D}^{-1/2}\mathbf{AD}^{-1/2}$ and $\mathbf{L}=\mathbf{I}_n-\mathbf{D}^{-1/2}\mathbf{AD}^{-1/2}$, respectively.
Note that the normalized graph Laplacian $\mathbf{L}$ is symmetric and positive semi-definite.
We express the eigenvalue decomposition of $\mathbf{L}$ as $\mathbf{U\Lambda U^\top}$, where $\mathbf{U}$ is a unitary matrix containing the eigenvectors, and $\mathbf{\Lambda}=\text{diag}\{\lambda_1, \lambda_2, \dots,\lambda_n\}$ comprises the eigenvalues of $\mathbf{L}$.
We usually modify the Laplacian with $\widehat{\mathbf{L}}=\frac{2\mathbf{L}}{\lambda_{\max}}-\mathbf{I}$ to scale the eigenvalues to $[-1,1]$. 
Note that the $\lambda_{\max}$ is set to $2$ in practice, then we have $\widehat{\mathbf{L}}\approx -\mathbf{P}$.

\header{\bf Spectral GNNs.} 
% A frequently utilized paradigm in GNNs can be defined as follows:
% \begin{equation}
% \label{Eqn:SpecFilter}
% \mathbf{X^{(\ell+1)}}=\sigma\left(\mathbf{\widetilde{P}}\mathbf{X}^{(\ell)}\mathbf{W}^{(\ell)}\right),
% \end{equation}
% where $\mathbf{X}^{(\ell)}$ corresponds to the node embeddings at the $\ell$-th layer, $\mathbf{W}^{(\ell)}$ is the learnable parameter matrix, and $\sigma(\cdot)$ represents the non-linear activation function, commonly chosen as $\text{ReLU}(\cdot)$.
%
Spectral-based GNNs exploit the spectral attributes of $G$ and apply the graph convolution operation on the spectral domain.
Many works~\citep{DBLP:APPNP, DBLP:ChebNetII} either approximate the filter with polynomial or exhibit similar properties of polynomial filters.
The graph filtering operation with respect to the graph Laplacian matrix $\mathbf{L}$ and signal $\mathbf{x}$ is defined as 
\begin{equation}
    \label{Eqn:SpecFilter}
    h(\mathbf{L})*\mathbf{x}
    =\mathbf{U}h(\mathbf{\Lambda})\mathbf{U}^\top\mathbf{x}
    \approx\mathbf{U}\left(\sum_{k=0}^K w_k\mathbf{\Lambda}^k\right)\mathbf{U}^\top\mathbf{x}
    =\left(\sum_{k=0}^K w_k\mathbf{L}^k\right)\mathbf{x},
\end{equation}
where $\mathbf{w}=[w_0,w_1,\dots,w_k]$ represents the polynomial coefficient vector. Given the graph filtering operation, the architecture of spectral GNNs is often expressed as
\begin{equation*}
    f(\mathbf{L},\mathbf{x})=f_{\theta_2}\left(h(\mathbf{L})*f_{\theta_1}(\mathbf{X})\right),
\end{equation*}
where $f_{\theta_i}(\cdot)$ represents the linear layer with coefficients $\theta_i$, and $\mathbf{X}$ combines multiple channels of signal $\mathbf{x}$.

As mentioned in Section~\ref{sec:intro}, the adoption of the detaching trick modifies the architecture of spectral GNNs to the following form:
\begin{equation*}
    f'(\mathbf{L},\mathbf{x})=f_{\theta_2}\left(f_{\theta_1}\left(h(\mathbf{L})*\mathbf{X}\right)\right).
\end{equation*}
The revised architecture differed from the original by excluding the passage of input features through a linear layer for dimensionality reduction.
Due to the absence of learnable parameters prior to graph propagation, the process of $\mathbf{L}^k\mathbf{X}$ can be precomputed.
Consequently, the learnable parameters control the linear combination and transformation of the propagated embeddings.
Given that the graph structure is engaged solely during graph propagation, the training process is inherently conductive to mini-batching.

\header{\bf Homophily.}
Homophily measures the tendency of the connected nodes on the graph to have the same label in node classification tasks.
This property heavily influences the classic GNN models which utilize one-hop neighbors, while the spectral GNNs can leverage multi-hop neighbor importance.
We usually quantify the homophily of a graph with the following edge homophily~\cite{DBLP:H2GCN}:
\begin{equation*}
\mathcal{H}(G)=\frac{1}{m}
\left| \left\{
(u,v):y(u)=y(v)\land(u,v)\in E
\right\} \right|,
\end{equation*}
where $y(\cdot)$ returns the label of nodes.
Intuitively, $\mathcal{H}(\cdot)$ denotes the ratio of homophilous edges on the graph.
A heterophilous graph implies $\mathcal{H}(G)\rightarrow 0$.

\section{Proposed Method}
\subsection{Motivation}

If we take retrospect on the spectral GNNs, this category of GNNs yields promising results, especially when applied to heterophilous datasets.
However, an aspect that has received less attention in existing spectral GNN research is scalability.
Numerous works\citep{DBLP:GraphSAGE, DBLP:GraphSAINT} that prioritize scalability resort to random sampling techniques to reduce the neighborhood of central nodes, or employ methods like historical embedding~\citep{DBLP:GNNAutoScale} to approximate node embeddings.
However, the convolution of spectral GNNs gathers information from up to $K$-hop neighbors, posing challenges to the effective deployment of scalable techniques.

Nevertheless, it is still possible to enhance the scalability of spectral GNNs using straightforward methods.
% Following SGC~\citep{DBLP:SGC}, many of the spectral methods~\citep{DBLP:ChebNetII, DBLP:OptBasis} decouple the graph propagation from training to simplify the network architecture.
% During the graph propagation, they compute $\mathbf{L}^k\mathbf{x}$ as described in Equation~\ref{Eqn:SpecFilter}, which can be preprocessed on the CPU.
As is mentioned in Section~\ref{sec:prelim}, the detaching trick may somewhat extend the scalability of spectral GNNs.
% This process is independent of the graph structure and can be mini-batched naturally.
% Nevertheless, this method introduces the following defects with scalability: 
% 1) Some researches~\citep{DBLP:SGCxGCN1, DBLP:SGCxGCN2} argue that separating training from the graph structure simplifies the network architecture and will negatively impact performance. 
% 2) The training is no longer end-to-end due to the graph propagation preprocess.
% Emerging approaches~\citep{DBLP:LM1, DBLP:LM2} based on language models~\citep{DBLP:BERT, DBLP:GPT3} enhance the performance on Ogbn-papers100M, which requires the raw text as input and end-to-end training.
% 3) Preprocess is potentially impractical when dealing with large graphs with high-dimensional node features since the dimension of original features cannot be reduced by MLP.
However, this trick is not without its drawbacks.
First, the decoupling of the GNNs leads to a non-end-to-end model.
Emerging approaches~\citep{DBLP:LM1, DBLP:LM2} based on language models~\citep{DBLP:BERT, DBLP:GPT3} enhance the performance on Ogbn-papers100M, which requires the raw text as input and end-to-end training.
% We cannot use the learnable language models to enhance the node representations with the raw text of large-scale datasets like Ogbn-papers100M.
Second, we cannot apply a learnable linear layer to reduce the dimension of raw features.
Preprocess is potentially impractical when dealing with large graphs with extremely high-dimensional node features.
Third, some researches~\citep{DBLP:SGCxGCN1, DBLP:SGCxGCN2} argue that separating training from the graph structure simplifies the network architecture and will negatively impact performance. 
Given these considerations, we propose ``Spectral Graph Neural Networks with Laplacian Sparsification (SGNN-LS)'' to enhance the scalability of spectral GNNs without decoupling the network architecture.

\subsection{Simplify the Graph Propagation with Laplacian Sparsification}

When we revisit Equation~\ref{Eqn:SpecFilter}, we observe that the filter step can be reformulated as a matrix multiplication involving the combined powers of Laplacian (referred to as $\mathbf{L}_K=\sum_{k=0}^K w_k\mathbf{L}^k$, Laplacian polynomial) and the input signal.
If we obtain the Laplacian polynomial $\mathbf{L}_K$, the propagation is then squeezed to a single step instead of multi-hop message-passing.
Since the computation and storage of matrix $\mathbf{L}_k$ is overwhelming, we introduce the Laplacian sparsification~\citep{DBLP:LS} to approximate the matrix within high probability and tolerable error.

Laplacian sparsification is designed to create a sparse graph that retains the spectral properties of the original graph.
In essence, the constructed graph, with fewer edges, is spectrally similar to the original one.
We provide a formal definition of the spectrally similar as follows.

\begin{definition}
    Given an weighted, undirected graph $G=(V,E,w)$ and its Laplacian $\mathbf{L}_G$. We say graph $G'$ is spectrally similar to $G$ with approximation error $\varepsilon$ if we have
\begin{equation*}
    (1-\varepsilon)\mathbf{L}_{G'}\preccurlyeq\mathbf{L}_G\preccurlyeq(1+\varepsilon)\mathbf{L}_{G'},
\end{equation*}
    where we declare that two matrix $\mathbf{X}$ and $\mathbf{Y}$ satisfy $\mathbf{X}\preccurlyeq\mathbf{Y}$ if $\mathbf{Y}-\mathbf{X}$ is positive semi-definite.
\end{definition}

A well-established result~\cite{matrix_analysis} states that $\mathbf{X}\preccurlyeq\mathbf{Y}$ implies $\lambda_i(\mathbf{X})\le\lambda_i(\mathbf{Y})$ for each $1\le i \le n$, where $\lambda_i(\cdot)$ denotes the $i$th largest eigenvalue of the matrix.
This corollary reveals that when two matrices, such as $\mathbf{L}_G$ and $\mathbf{L}_{G'}$, exhibit spectral similarity, their quadratic form and eigenvalues are in close correspondence.
Graph sparsification is a fundamental problem of graph theory.
Many studies~\citep{DBLP:sparsifier} have introduced algorithms that generate $\varepsilon$-sparsifiers for a given graph.

The sole difference between scaled Laplacian and the propagation matrix is a negative sign. We have

\begin{equation}
    \label{Eqn:LKver-rw}
    \mathbf{L}_K 
    = \sum_{k=0}^K w_k\mathbf{L}^K
    \approx \sum_{k=0}^K w_k\mathbf{P}^k 
    = \mathbf{D}^{-1/2}\cdot\mathbf{D}\left(\sum_{k=0}^K w_k\left(\mathbf{D}^{-1}\mathbf{A}\right)^k \right)\mathbf{D}^{-1/2},
\end{equation}

where the negative sign can incorporated into the coefficients $w_k$. 
From Equation~\ref{Eqn:LKver-rw}, we observe that we may convert our desiring matrix $\mathbf{L}_K$ to a random walk matrix polynomial $\left(\mathbf{D}^{-1}\mathbf{A}\right)^k$ with coefficients $[w_0,...,w_K]$.
The prior works~\citep{DBLP:sparsifierER, DBLP:sparsifierRWP} has presented a promising construction of the sparsifier of a given graph, whose pseudo-code is presented in Algorithm~\ref{alg:edgesamp_ER}.
Following them, we have a promising guarantee to approximate random walking matrix polynomial by effective resistance termed Theorem~\ref{theorem:rwps}.

\begin{theorem}
    \label{theorem:rwps}
    \textbf{(Random walk polynomial sparsification.)} For any unweighted, undirected graph $G$ with $n$ vertices and $m$ edges, any $\mathbf{w}=[w_0, w_1, ..., w_K]\in\mathcal{R}_+^{(K+1)}, \mathbf{w}\ne\mathbf{0}$ and any approximation parameter $\varepsilon$, we can construct an unbiased $\varepsilon$-sparsifier of the random walk matrix polynomial $\sum_{k=0}^K w_k\mathbf{D}\left(\mathbf{D}^{-1}\mathbf{A}\right)^k$ within $O(n\log n/\varepsilon^2)$ edges and probability $1-K/n$ at least.
\end{theorem}

We have extended the original theorem proposed by~\citep{DBLP:sparsifierRWP} to accommodate non-normalized polynomial coefficients $\mathbf{w}$.
Sampling an edge from $\mathbf{D}(\mathbf{D^{-1}}\mathbf{A})^k$ is an atomic operation in constructing our desired graph, and it is frequently employed in our subsequent algorithms.
We state this procedure in Algorithm~\ref{alg:edgesamp}.
In the upcoming sections, we will present our comprehensive algorithms, which include the weight correction and the adaptation for both static and learnable polynomial coefficients $\mathbf{w}$.
For in-depth discussion regarding the approximation estimation, correctness proof, and complexity analysis, please retrieve section~\ref{sec:theorem} and appendix~\ref{app:theorem}.

\begin{figure*}[t]
  \centering
  \includegraphics[width=0.8\textwidth]{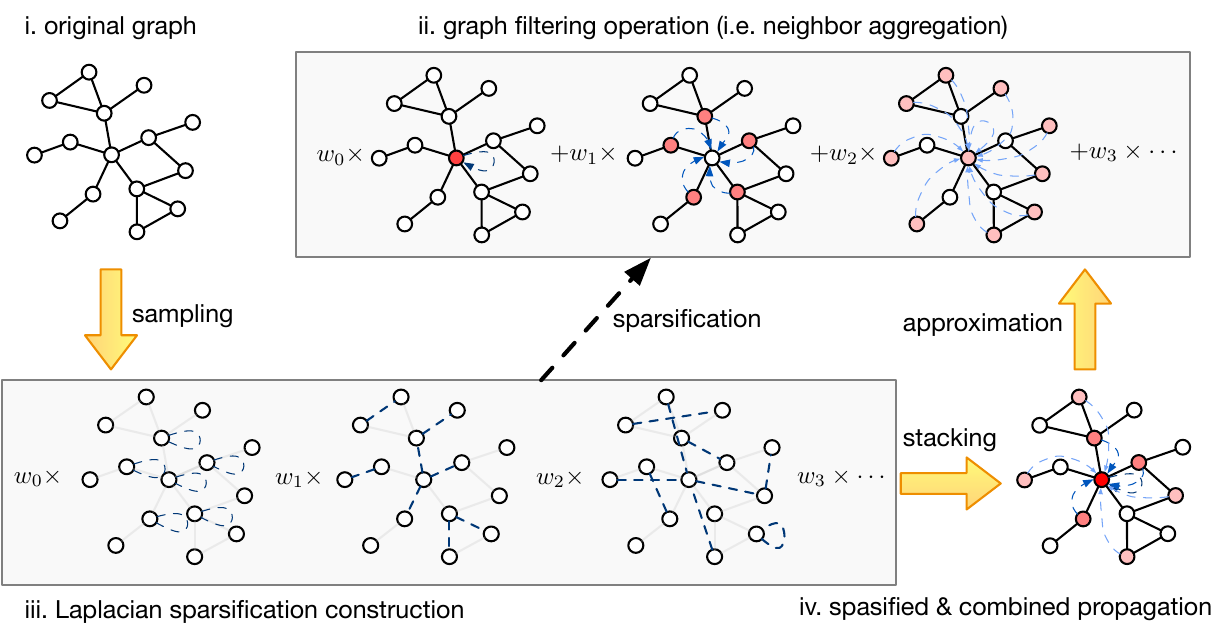}
  \caption{An overview of how Laplacian sparsification works. For clarity, the propagation of one single center node is illustrated. Laplacian sparsification is applied to the entire graph, generating fully sparsified graphs to satisfy the propagation requirements of all the nodes in the graph.}
  \label{fig:overview}
\end{figure*}

\begin{algorithm}[t]
    \DontPrintSemicolon
    \caption{{\sf Edge Sampling of $\mathbf{D}(\mathbf{D^{-1}}\mathbf{A})^k$}}\label{alg:edgesamp}
    \KwIn{Edge set $E$, power index $k$.}
    \KwOut{Sampled edge $(u,v)$}
    $e \gets$ sample an edge from $E$ uniformly at random\;
    $i \gets$ sample an integer in $[0,k-1]$ uniformly at random\;
    $u \gets $ the end of random walk on $E$ (i.e. graph $G$), starting from $e_u$ with length $i$\;
    $v \gets $ the end of random walk on $E$ (i.e. graph $G$), starting from $e_v$ with length $k-i-1$\;
    \Return $(u, v)$
\end{algorithm}

\subsubsection{Laplacian Sparsification for Static Polynomial Coefficients}
Some of the early classic GNNs, like GCN~\citep{DBLP:GCN} and APPNP~\citep{DBLP:APPNP}, employ static Laplacian polynomials as the graph filter.
For example, the propagation layer of APPNP fixes the weight $w_k=\alpha(1-\alpha)^{k},k\ne K$, and $w_K=(1-\alpha)^K$ for different hops, where $\alpha$ is the restart probability of random walk.

Note that if we repeat Algorithm~\ref{alg:edgesamp} for $M$ times with the same power index $k$ and apply each obtained edge with the weight $m/M$, we can approximate $\mathbf{D}(\mathbf{D}^{-1}\mathbf{A})^k$.
This result is still distant from our desired form in Equation~\ref{Eqn:LKver-rw}.
First, we need to approximate the random walk matrix polynomial based on our existing method of approximating a term $\mathbf{D}(\mathbf{D}^{-1}\mathbf{A})^k$.
One intuitive and efficient idea is to distribute the $M$ edges among all $K+1$ subgraphs of $\mathbf{D}(\mathbf{D}^{-1}\mathbf{A})^k$.
The number of edges assigned to each subgraph follows a multinomial distribution with weights $\mathbf{w}$. 
Note that $w_k$ is not guaranteed to be positive.
The absolute value of $w_k$ is proportional to the probability of being sampled, while $\text{sgn}(w_k)$ decides the sign of the edge weight.
We select a random walk length $k$ based on the probability distribution $\text{Pr}\{k=i\}=\vert w_i\vert/\Vert\mathbf{w}\Vert_1$.
Second, we execute Algorithm~\ref{alg:edgesamp} with edge set $E$ and power index $k$ to generate an edge $(u,v)$.
Given that we are now approximating the graph $\mathbf{D}(\mathbf{D}^{-1}\mathbf{A})^k$, the edge value is adjusted to 
\begin{equation*}
    \text{sgn}(w_k)\Vert\mathbf{w}\Vert_1\cdot d_u^{-1/2}d_v^{-1/2}\cdot m/M.
\end{equation*}

We provide the pseudo-code for constructing a Laplacian sparsified random walk matrix polynomial with static coefficients in Appendix~\ref{app:APPNP-LS}. Additionally, we offer an example of such a model integrated with our method APPNP-LS in Appendix~\ref{app:APPNP-LS}.

\subsubsection{Laplacian Sparsification for Learnable Polynomial Coefficients}

Several recent spectral works, like GPR-GNN~\citep{DBLP:GPR-GNN}, BernNet~\citep{DBLP:recommendationsystem1}, and ChebNetII~\citep{DBLP:ChebNetII}, employ learnable polynomial coefficients to dynamically adapt the proper filter.
For example, GPR-GNN uses monomial bases to identify the optimal filter, while BernNet employs Bernstein polynomial basis to align with the property that the eigenvalues of normalized Laplacian fall within the range $[0,2]$.

Upon revisiting the procedure stated above, it becomes apparent that the polynomial coefficients $\mathbf{w}$ primarily affect the sampling of the power index $k$ and the adjustment of edge weights.
To facilitate the training of $\mathbf{w}$, we need to calculate the derivative of each $w_k$ for gradient descent.
However, we cannot obtain the correct derivatives of $\mathbf{w}$ since all the $w_k$ equally contribute to the part that generates gradients.
To address this issue, we directly multiply the polynomial coefficients $\mathbf{w}$ with the weight of the sampled edges instead of the sampling with probability $|w_k|/\Vert\mathbf{w}\Vert_1$ in SLSGC.
Thus, the edge weight becomes $w_kd_u^{-1/2}d_v^{-1/2}\cdot m/M$.
This adjustment connects the gradient of $\mathbf{w}$ with the message passed by the corresponding edges, ensuring the correct derivative chain for training $\mathbf{w}$.
However, this modification splits coefficients $w_k$, leading to independent sampling for each hop $k$ and potentially sacrificing the efficiency.
Theoretically, we need to sample more edges to support the training of $\mathbf{w}$ while maintaining the bound of approximation.

We provide the pseudo-code of constructing a Laplacian sparsified random walk matrix polynomial with learnable coefficients (named GLSGC) in Appendix~\ref{app:GPR-LS}.
%which is similar to Algorithm~\ref{alg:APPNP-LS} and more straightforward. 
Besides, we offer an example of such a model integrated with our method GPR-LS in Appendix~\ref{app:GPR-LS}

\subsection{Node-Wise Laplacian Sampling for Semi-supervised Tasks}
\label{sec:semi}
For representation learning tasks on large-scale graphs with few training nodes, classic S/GLSGC are wasteful as most edges are sampled between nodes in the validation/test set.
To address this issue, we propose a node-wise sampling method to approximate the corresponding rows of the result of Equation~\ref{Eqn:LKver-rw} for the training nodes.
This enhancement significantly improves the training efficiency, as nodes in the validation/test set will not aggregate information during training.

Reflecting on Equation~\ref{Eqn:LKver-rw}, it becomes clear that $((\mathbf{D}^{-1}\mathbf{A})^k)_{i,j}$ determines the probability of a random walk of length $k$ starting from node $i$ and ending at node $j$.
This inspires us that we can ensure at least on incident node of the sampled edge belongs to the training set by directly sampling the random walk from the training set.
Following Equation~\ref{Eqn:LKver-rw}, for each random walk length $k$, we first distribute the $M$ sampled edges among all nodes in proportion to their degrees.
Then we perform random walk samplings and correct the value of each generated edge $(u_0, u_k)$ to $d_{u_0}^{-1/2}d_{v_0}^{-1/2}\cdot w_k/M$ for each sampled walk $(u_0,u_1,...,u_k)$.

It can be mathematically proven that the proposed algorithm produces an unbiased approximation of the selected rows of $\mathbf{D}(\mathbf{D}^{-1}\mathbf{A})^k$.
Moreover, this method is node-wise, allowing for natural mini-batching.
As a result, we achieve a strong scalability promotion of the spectral methods.
The pesudo-code of this algorithm is proposed in Appendix~\ref{app:semi}

\section{Theoretical Analysis}
\label{sec:theorem}
In this section, we will conduct a comprehensive analysis, including rigorous proofs of correctness, complexity, and other key properties of the methods we have introduced.
Due to the space limitation, time and space complexity analysis is proposed in Appendix~\ref{app:overheadanalysis}

\subsection{Error Guarantee about Laplacian Sparsification}

From a graph theory perspective, a weighted graph is closely related to electric flow.
Each edge with weight $w(e),e=(u,v)\in E$ ($w(e)=1$ for unweighted graphs) can be considered equivalent to a resistor with resistance $1/w(e)$ connecting nodes $u$ and $v$.
When we view the graph as a large complex resistor network, the resistance between $u$ and $v$ is defined as the effective resistance $R(u,v)$.

\begin{algorithm}[t]
    \DontPrintSemicolon
    \caption{{\sf Edge Sampling by Effective Resistance}}
    \label{alg:edgesamp_ER}
    \KwIn{Edge set $E$, upper bound of effective resistance $R_{\text{sup}}$, sampling number $M$.}
    \KwOut{Sampled edge set $\widetilde{E}$}
    $\widetilde{E} \gets \emptyset$   \\ 
    \For {$i$ {\rm from} $1$ {\rm to} $M$}{
        $\widetilde{E} \gets \widetilde{E}\ \cup$ sampled edge $e$ with probability $p(e)\varpropto w(e)R_{\text{sup}}(e)$ and weight $1/(M\cdot R_{\text{sup}}(e))$
    }
    \Return $E$
\end{algorithm}

\begin{lemma}
    \label{lem:ERupperbound}
    ~\citep{DBLP:sparsifierRWP} \textbf{(Upper bound of effective resistance.)}
    The effective resistance between two nodes $u$ and $v$ on graph $G_r$ is upper bounded by 
    \vspace{-0.5em}
    \begin{equation*}
        R_{G_r}(u,v)\le \sum_{j=0}^{r-1}\frac{2}{\mathbf{A}(i_{j-1},i_j)}=R_{\text{sup},G_r}(u,v),
    \end{equation*}
    \vspace{-0.5em}
    where $(u=i_0, i_1, \cdots, i_{r-1}, v=i_r)$ is a path on $G$.
\end{lemma}

% This is a general version of the upper bound of effective resistance about the graph $G_r$.
As we are primarily concerned with unweighted graph $G$, the upper bound can be simplified to a constant $2r$.
We can prove a more robust conclusion that Algorithm~\ref{alg:edgesamp} draws path $p$ on graph $G$ with the probability \textbf{strictly} proportional to $w(p)$. 
This probability is independent of the $R_{\text{sup}, G_r}(\cdot)$, which means that running a Monte-Carlo sampling on the graph yields an \textbf{unbiased approximation} of $G_r$.
% Algorithm~\ref{alg:edgesamp} represents a standard procedure of effective resistance based graph sampling.
Hence, by replacing the sampling process in Algorithm~\ref{alg:edgesamp_ER} with Algorithm~\ref{alg:edgesamp}, we obtain an unbiased graph sparsifier generator of $G_r$ with $O(rm\log n/\varepsilon^2)$ edge.
This sparsifier can be further reduced to $O(n\log n/\varepsilon^2)$ by the existing works~\citep{DBLP:LS, DBLP:sparsifierER}.
% Both our proposed methods, SLSGC and GLSGC, follow the complexity outlined in Theorem~\ref{theorem:rwps}.
The proof of Theorem~\ref{theorem:rwps} is detailed in Appendix~\ref{app:theorem}.
% In practice, the sampling number can be set rather smaller than the theoretical bound to achieve the desired performance.

In the method proposed in Section~\ref{sec:semi}, the sampled random walk of length $r$ originating from a distinct source node $u$ also shares the same upper bound of effective resistance $2r$, which can be equivalently considered as the effective resistance based Laplacian sparsification.
Moreover, our method is more intuitive and simplified since we directly sample the desired edge proportional to $w(p)$, without relying on effective resistance.
We take consideration of the single candidate start $u$ for the random walks and define $c_r(u,v)$ as the final generated edge weight of $(u,v)$ on graph $G_r$.

\begin{theorem}
    \label{the:nodewise}
    Given a weighted graph $G$, and the candidate set $U$ of the random walk starts. For any $u\in U$ and $v\in V$, we have $c_r(u,v)$ is the unbiased approximation of $\left(\mathbf{D}(\mathbf{D}^{-1}
    \mathbf{A})^r\right)_{u,v}$.
\end{theorem}

The detailed proof is presented in Appendix~\ref{app:theorem}.
In practice, we do not need many samples to achieve superior performance for all our proposed methods.
This node-centered sampling method enables us to adapt our method to semi-supervised tasks, saving the memory for a larger batch size.

\subsection{Error Guarantee about Propagated Signals}
Having established the similarity between the original propagation matrix and its approximation, a discernible gap between graph theory and machine learning remains.
This section delves deeper into the variances between the propagate signals when utilizing either the original or approximated propagation matrices.

We exemplify this investigation with the APPNP model.
Given that different signal channels remain independent during propagation, it is feasible to analyze each channel individually w.l.o.g.. 
Denote $\mathbf{z}^{(t)}$ as the output signal of APPNP following $t$ rounds of propagation.
It has been shown that $\mathbf{z}^{(t)}$ converges to the solution of the following linear system when $t\rightarrow\infty$:
\begin{equation}
    \label{eqn:appnpprop}
    \left(\mathbf{I}-(1-\alpha)\mathbf{D}^{-1/2}\mathbf{AD}^{-1/2}\right)\mathbf{z}=\alpha \mathbf{x},
\end{equation}
where $\mathbf{x}$ is the input signal vector.
As is investigated in~\cite{DBLP:APPNPloss}, the solution to such a linear system is the optima of some convex quadratic optimization problem.

\begin{proposition}
    ~\citep{DBLP:APPNPloss} Let $\mathbf{z}^*$ be the optima of the following optimization problem.
    \begin{equation}
    \label{eqn:appnploss}
        \min_{\mathbf{z}\in\mathbb{R}^n}(1-\alpha)\mathrm{Tr}(\mathbf{z}^\top \mathbf{L}\mathbf{z})+\alpha\Vert{\mathbf{z}-\mathbf{x}}\Vert_F^2
    \end{equation}
    Then $\mathbf{z}^*$ is the unique solution to the Equation~\ref{eqn:appnpprop}.
\end{proposition}

To qualify the similarity between any determined $\mathbf{z}$ and $\mathbf{z}^{(\infty)}$, we employ the loss function$\mathcal{L}(\mathbf{z})=(1-\alpha)\text{Tr}(\mathbf{z}^\top \mathbf{L}\mathbf{z})+\alpha\Vert{\mathbf{z}-\mathbf{x}}\Vert_F^2$.
Furthermore, we aim to demonstrate that the difference in loss, calculated for propagated signals under different propagation matrices is theoretically bounded.

\begin{theorem}
    \label{the:appnplosserror}
    For any graph $G$, given input signal $\mathbf{x}$, the propagated signal $\mathbf{z}$ after $K$ rounds of APPNP propagation, and the propagated $\mathbf{\tilde{z}}$ after one round of propagation with an $\varepsilon$-sparsifier of corresponding random walk matrix polynomial. The relative error of the loss function $\mathcal{L}(\mathbf{z})=(1-\alpha)\mathrm{Tr}(\mathbf{z}^\top \mathbf{L}\mathbf{z})+\alpha\Vert{\mathbf{z}-\mathbf{x}}\Vert_F^2$ is
    \begin{equation*}
        \left\Vert\frac{\mathcal{L}(z)-\mathcal{L}(\tilde{z})}{\mathcal{L}(z)}\right\Vert=O(\varepsilon).
    \end{equation*}
\end{theorem}

The comprehensive formal proof of Theorem~\ref{the:appnplosserror} is detailed in Appendix~\ref{app:theorem}.
The Theorem~\ref{the:appnplosserror} shows that the approximated propagation matrix conserves the core attributes of the propagated signal, thereby preserving the fundamental aspects of the model.

\section{Related Works}
\header{\bf Spectral GNNs.}
To align with the proposed methods above, we categorize the spectral GNNs based on their used polynomial filter: static (predefined) polynomial filters and learnable polynomial filters.
For static polynomial filters, GCN~\citep{DBLP:GCN} uses a fixed simplified Chebyshev polynomial approximation and operates as a low-pass filter.
APPNP~\citep{DBLP:APPNP} combines the GCN with Personalized PageRank, which can approximate more types of filters but still cannot operate as an arbitrary filter.
GNN-LF/HF~\citep{DBLP:GNN-LF/HF} predefines the graph filter from the perspective of graph optimization.
For learnable polynomial filters, ChebNet~\citep{DBLP:ChebNet} first approximates the desired filter with the Chebyshev polynomial base, which can operate as an arbitrary filter theoretically.
Similarly, GPRGNN~\citep{DBLP:GPR-GNN} considers the monomial base to learn the importance of $k-$hop neighbors directly.
BernNet~\citep{DBLP:BernNet} uses the Bernstein polynomial base to make the filter semi-positive definite.
ChebNetII~\citep{DBLP:ChebNetII} revisits the ChebNet and proposes a filter design via Chebyshev interpolation.
~\cite{DBLP:OptBasis} proposes FavardGNN to learn basis from all possible orthonormal bases and OptBasisGNN to compute the best basis for the given graph.

\header{\bf Subsampling-based Scalable GNNs.}
Many subsampling~\citep{DBLP:LADIES, DBLP:SIGN} methods were studied when the training of GNNs faced the memory limit.
This strategy can be broadly divided into two sorts: node sampling methods and subgraph sampling methods.
GraphSAGE~\citep{DBLP:GraphSAGE} randomly samples the neighborhood of the aggregation center to approximate the graph propagation.
Instead of node-centered sampling, FastGCN~\citep{DBLP:FastGCN} deploys a layer-wise sampling method to limit the upper bound of graph propagation in each network layer.
Besides, METIS, a well-known clustering algorithm, is used in generating mini-batched graph data for the training of Cluster-GCN~\citep{DBLP:ClusterGCN}, GAS~\citep{DBLP:GNNAutoScale}, LazyGNN~\citep{DBLP:LazyGNN}, and LMC-GCN~\citep{DBLP:LMC}.
GraphSAINT~\citep{DBLP:GraphSAINT} invents multiple new subgraph sampling methods and corresponding normalization coefficients for unbiased training.

Node sampling methods are mostly designed for simplifying and approximating the propagation of the vanilla GCN, and the subgraph sampling methods hinder the long-range interaction between the nodes. 
These methods are not suitable for extending the scalability of the spectral GNNs discussed in our paper.

\header{\bf Graph Sparsification.} 
Graph sparsification is a consistently studied topic in graph theory.
~\cite{DBLP:sparsifyintro} introduces the concept of graph sparsification and presents an efficient algorithm to approximate the given graph Laplacian with a smaller subset of edges while maintaining the spectral properties.
~\cite{DBLP:sparsifierER} leverages effective resistance to yield the promising random sampled edges on the graph.
~\cite{DBLP:sparsifierLinearsized} proposes the first method to construct linear-sized spectral sparsification within almost linear time.
Our work is mainly enlightened by the works~\citep{DBLP:NetSMF} and~\citep{DBLP:sparsifierRWP}, which make an approximation to the series of the multi-hop random walk sampling matrix.

In the context of GNNs, many works like~\citep{DBLP:sparGNN1, DBLP:sparGNN2} involve graph sparsification to enhance efficiency or performance.
As they usually apply sparsification to the original graph, these methods are not promising in the scenario of multi-step graph propagation which is frequently used in spectral GNNs.

\section{Experiments}
%In this section, we conduct an extensive experiment and make a thorough analysis to validate the effectiveness and scalability of our proposed GNN-LS. 

In this section, we will first describe our experimental settings.
Next, we will conduct a comprehensive analysis of the experimental results.
Due to the space limitation, we provide additional experimental results in Appendix~\ref{app:expres}, including the applicability of our methods on multilayer models, the comparison with ClusterGCN and H2GCN, the comparison among detached models, and some analysis on time, space, executability, and sampling numbers.

\begin{table*}[t]
\caption{Experimental results of some baselines and our Laplacian sparsification entangled methods on multiple \textbf{small-scale datasets}. Model name with suffix ``-LS'' represents the spectral methods entangled with our proposed method. The line beginning with ``Avg. $\Delta$'' reveals the average performance difference between the base model and its LS-variation. Most evaluation metrics are accuracy(\%), but ROC AUC (\%) for datasets with 2 classes (Twitch-de, and Twitch-gamers / Penn94 in Table~\ref{tbl:res-large}). The \textbf{bold} font highlights the best results, whereas the \underline{underlined} numbers indicate the second and third best.}
\label{tbl:res-small}
\resizebox{\linewidth}{!}{
\centering

\begin{tabular}{lcccccccccc}
\toprule
 Dataset            & Cora & Citeseer & PubMed & Actor & Wisconsin 
                    & Cornell & Texas & Photo & Computers & Twitch-de \\
$\mathcal{H}(G)$    & 0.810 & 0.736 & 0.802 & 0.219 & 0.196 
                    & 0.305 & 0.108 & 0.827 & 0.777 & 0.632 \\
\midrule
MLP                 & 76.72$_{\pm0.89}$ & 77.29$_{\pm0.32}$ & 86.48$_{\pm0.20}$ & 39.99$_{\pm0.76}$ & 90.75$_{\pm2.38}$ 
                    & \underline{92.13}$_{\pm1.80}$ & \underline{92.13}$_{\pm1.64}$ & 90.11$_{\pm0.33}$ & 85.00$_{\pm0.35}$ & 68.84$_{\pm0.54}$ \\
GAT                 & 86.80$_{\pm0.94}$ & 81.16$_{\pm0.97}$ & 86.61$_{\pm0.35}$ & 35.26$_{\pm0.82}$ & 69.13$_{\pm3.00}$ 
                    & 78.36$_{\pm1.80}$ & 79.02$_{\pm2.95}$ & 93.31$_{\pm0.34}$ & 88.39$_{\pm0.35}$ & 67.90$_{\pm0.75}$ \\
GCNII               & 88.52$_{\pm1.03}$ & 81.24$_{\pm0.65}$ & 89.17$_{\pm0.40}$ & 41.20$_{\pm0.82}$ & 82.88$_{\pm2.50}$
                    & 90.49$_{\pm1.64}$ & 84.75$_{\pm3.44}$ & 94.20$_{\pm0.23}$ & 88.55$_{\pm0.61}$ & 68.03$_{\pm0.33}$ \\
PPRGo               & 87.37$_{\pm0.95}$ & 80.76$_{\pm0.52}$ & 88.35$_{\pm0.34}$ & 39.96$_{\pm0.25}$ & \underline{93.13}$_{\pm1.63}$ 
                    & 90.49$_{\pm3.28}$ & 89.67$_{\pm1.80}$ & 93.73$_{\pm1.80}$ & 87.20$_{\pm0.34}$ & 71.01$_{\pm0.61}$ \\
LMCGCN              & 86.67$_{\pm1.12}$ & 77.58$_{\pm0.73}$ & 89.86$_{\pm0.18}$ & 35.20$_{\pm1.43}$ & 70.25$_{\pm3.63}$
                    & 78.36$_{\pm1.97}$ & 79.84$_{\pm1.97}$ & 94.12$_{\pm0.44}$ & \underline{90.67}$_{\pm0.30}$ & 68.29$_{\pm0.76}$\\
LazyGNN             & \underline{89.23}$_{\pm0.71}$ & 79.37$_{\pm1.02}$ & 89.67$_{\pm0.56}$ & 40.94$_{\pm0.80}$ & 91.13$_{\pm1.88}$
                    & 86.56$_{\pm1.64}$ & 87.21$_{\pm3.44}$ & 95.10$_{\pm0.27}$ & 90.54$_{\pm0.31}$ & 67.44$_{\pm0.59}$\\
\midrule
GCN                 & 87.78$_{\pm1.05}$ & 81.50$_{\pm0.93}$ & 87.39$_{\pm0.42}$ & 35.62$_{\pm0.52}$ & 65.75$_{\pm3.00}$
                    & 71.96$_{\pm7.86}$ & 77.38$_{\pm1.97}$ & 93.62$_{\pm0.35}$ & 88.98$_{\pm0.37}$ & \underline{73.72}$_{\pm0.61}$ \\
SGC                 & 87.24$_{\pm0.97}$ & 81.53$_{\pm0.87}$ & 87.17$_{\pm0.15}$ & 34.40$_{\pm0.58}$ & 67.38$_{\pm3.50}$
                    & 70.82$_{\pm7.70}$ & 79.84$_{\pm1.64}$ & 93.41$_{\pm0.35}$ & 88.61$_{\pm0.30}$ & \underline{73.70}$_{\pm0.67}$ \\
% $\Delta$            & -0.54 & +0.03 & -0.22 & -1.22 & +1.63 
%                     & -1.14 & +2.46 & -0.21 & -0.37 & -0.02 \\
\midrule
GPR                 & 88.80$_{\pm1.17}$ & \underline{81.57}$_{\pm0.82}$ & \textbf{90.98}$_{\pm0.25}$ & 40.55$_{\pm0.96}$ & 91.88$_{\pm2.00}$ 
                    & 89.84$_{\pm1.80}$ & \textbf{92.78}$_{\pm2.30}$ & 95.10$_{\pm0.26}$ & 89.69$_{\pm0.41}$ & \textbf{73.91}$_{\pm0.65}$ \\
GPR-LS              & \textbf{89.31}$_{\pm1.07}$ & \underline{81.65}$_{\pm0.53}$ & \underline{90.95}$_{\pm0.37}$ & \underline{41.82}$_{\pm0.55}$ & 93.63$_{\pm2.88}$ 
                    & \underline{91.15}$_{\pm1.15}$ & \underline{92.62}$_{\pm1.48}$ & \underline{95.30}$_{\pm0.22}$ & 90.47$_{\pm0.41}$ & 73.49$_{\pm0.51}$ \\
% $\Delta$            & +0.51 & +0.08 & -0.03 & +1.27 & +1.75 
%                     & +1.30 & -0.16 & +0.20 & +0.78 & -0.42 \\
\midrule
Jacobi              & 88.46$_{\pm0.93}$ & 80.22$_{\pm0.61}$ & \underline{90.21}$_{\pm0.44}$ & $41.03_{\pm0.94}$ & 89.38$_{\pm3.26}$ 
                    & 89.18$_{\pm2.95}$ & 89.02$_{\pm3.44}$ & 94.33$_{\pm3.44}$ & 89.77$_{\pm0.38}$ & 69.57$_{\pm2.15}$ \\
Jacobi-LS           & \underline{89.23}$_{\pm0.74}$ & 81.43$_{\pm0.74}$ & 89.87$_{\pm0.44}$ & $41.12_{\pm0.80}$ & \textbf{93.75}$_{\pm2.63}$ 
                    & 89.67$_{\pm2.30}$ & 90.66$_{\pm2.30}$ & \textbf{95.37}$_{\pm2.30}$ & \underline{90.76}$_{\pm0.31}$ & 73.29$_{\pm0.82}$\\
% $\Delta$            & +0.77 & +1.21 & -0.34 & +0.09 & +4.37 
%                     & +0.49 & +1.64 & +1.04 & +0.99 & +3.72\\
\midrule
Favard              & 86.65$_{\pm1.00}$ & 81.13$_{\pm0.86}$ & 89.87$_{\pm0.30}$ & \underline{41.39}$_{\pm0.53}$ & 92.25$_{\pm2.25}$ 
                    & 86.72$_{\pm2.79}$ & 90.00$_{\pm1.97}$ & 94.35$_{\pm0.30}$ & 89.43$_{\pm0.29}$ & 72.78$_{\pm0.47}$\\
Favard-LS           & 88.65$_{\pm1.07}$ & 81.34$_{\pm0.68}$ & 90.13$_{\pm0.33}$ & $41.00_{\pm0.91}$ & \underline{92.50}$_{\pm2.13}$
                    & 86.56$_{\pm3.93}$ & 89.70$_{\pm3.77}$ & \underline{95.29}$_{\pm0.31}$ & \textbf{90.96}$_{\pm0.29}$ & 73.29$_{\pm0.76}$\\
% $\Delta$            & +2.00 & +0.21 & +0.26 & -0.39 & +0.25 
%                     & -0.16 & -0.30 & +0.94 & +1.53 & +0.51\\
\midrule
APPNP               & 88.69$_{\pm1.00}$ & 81.32$_{\pm0.68}$ & 88.49$_{\pm0.28}$ & 40.73$_{\pm0.67}$ & 90.38$_{\pm2.38}$ 
                    & 90.98$_{\pm2.13}$ & 90.82$_{\pm2.79}$ & 93.82$_{\pm0.26}$ & 86.97$_{\pm0.35}$ & 68.29$_{\pm0.72}$ \\
APPNP-LS            & 88.44$_{\pm1.10}$ & \textbf{82.28}$_{\pm0.49}$ & 88.70$_{\pm0.45}$ & \textbf{41.98}$_{\pm0.43}$ & 91.00$_{\pm3.13}$ 
                    & \textbf{92.30}$_{\pm0.98}$ & 90.98$_{\pm1.64}$ & 93.79$_{\pm0.36}$ & 87.84$_{\pm0.34}$ & 72.82$_{\pm0.46}$ \\
% $\Delta$            & -0.25 & +0.96 & +0.21 & +1.25 & +0.62 
%                     & +1.32 & +0.16 & -0.03 & +0.87 & +4.54 \\
\midrule
Avg. $\Delta$       & +0.76 & +0.62 & +0.03 & +0.56 & +1.44 & +0.74 & +0.34 & +0.54 & +1.04 & +2.09 \\
\bottomrule
\end{tabular}
}
\vspace{-0.4em}
\end{table*}

\subsection{Tested Models, Datasets, and Configurations}
\header{\bf Tested models.} We compare our method with the vanilla MLP, classic GNNs like GCN, GCNII~\citep{DBLP:GCNII}, and GAT~\citep{DBLP:GAT}, detaching methods like SGC, spectral GNN with static polynomial coefficients like APPNP, and spectral GNN with learnable coefficients GPR-GNN, JacobiConv, and FavardGNN.
Meanwhile, we involve LazyGNN and LMCGCN to show the ability of up-to-date scalable methods, and PPRGo~\citep{DBLP:PPRGo} to test the efficacy of our model when dealing with high-dimensional data.
For spectral GNNs, we entangle them with our proposed Laplacian sparsification method for the effectiveness test.
All the baselines are reimplemented with Pytorch~\citep{DBLP:Pytorch} and PyG~\citep{DBLP:PyG} library modules as competitors.
Codes, detailed parameter matrix, and reproduction guidance are stated in Appendix~\ref{app:params}.

\header{\bf Datasets.} 
All the baselines and our proposed method are tested with various datasets with diverse homophily and scales, including Cora, Citeseer, PubMed~\citep{DBLP:cora1, DBLP:cora2}, Photos, computers~\citep{DBLP:photo1, DBLP:photo2}, Actor~\citep{DBLP:Geom-GCN}, Cornell, Texas, Wisconsin~\citep{DBLP:Geom-GCN}, Twitch-de, Twitch-gamers~\citep{DBLP:LINKX},  Penn94~\citep{DBLP:LINKX}, Ogbn-arxiv, Ogbn-papers100M~\citep{DBLP:OGB} and MAG-scholar-C~\citep{DBLP:MAG}.
The downstream task of these datasets is node classification.
Detailed information on the datasets is stated in Appendix~\ref{app:dataset}.

\header{\bf Configurations.} 
All the experiments except those with dataset Ogbn-papers100M are conducted on the server equipping GPU NVIDIA A100 40GB.
For Ogbn-papers100M and MAG-scholar-C, we deploy our experiment on the server with GPU NVIDIA A100 80GB and 512G RAM.
Detailed information on experimental environments and package versions are stated in Appendix~\ref{app:config}.

\subsection{Accuracy of the Approximation}

Before analyzing the test results of SGNN models with and without our plug-in unit, we first examine the similarity between the precisely calculated filtered matrix ($\sum_{k=0}^K w_k\mathbf{P}^K$) and the approximated matrices.
Compared to other spectral methods, the only new hyperparameter introduced is ``-\noindent-ec'', which controls the sampling numbers of each propagation step to $\text{ec}\cdot n\log n$. 
Although it is not possible to directly calculate $\varepsilon$ from ``-\noindent-ec'' due to the theoretical use of Big-Oh notation, there is a clear inverse relationship: increasing ``-\noindent-ec'' decreases the approximation error $\varepsilon$.

\begin{figure}[h]
  \centering
  \includegraphics[width=\linewidth]{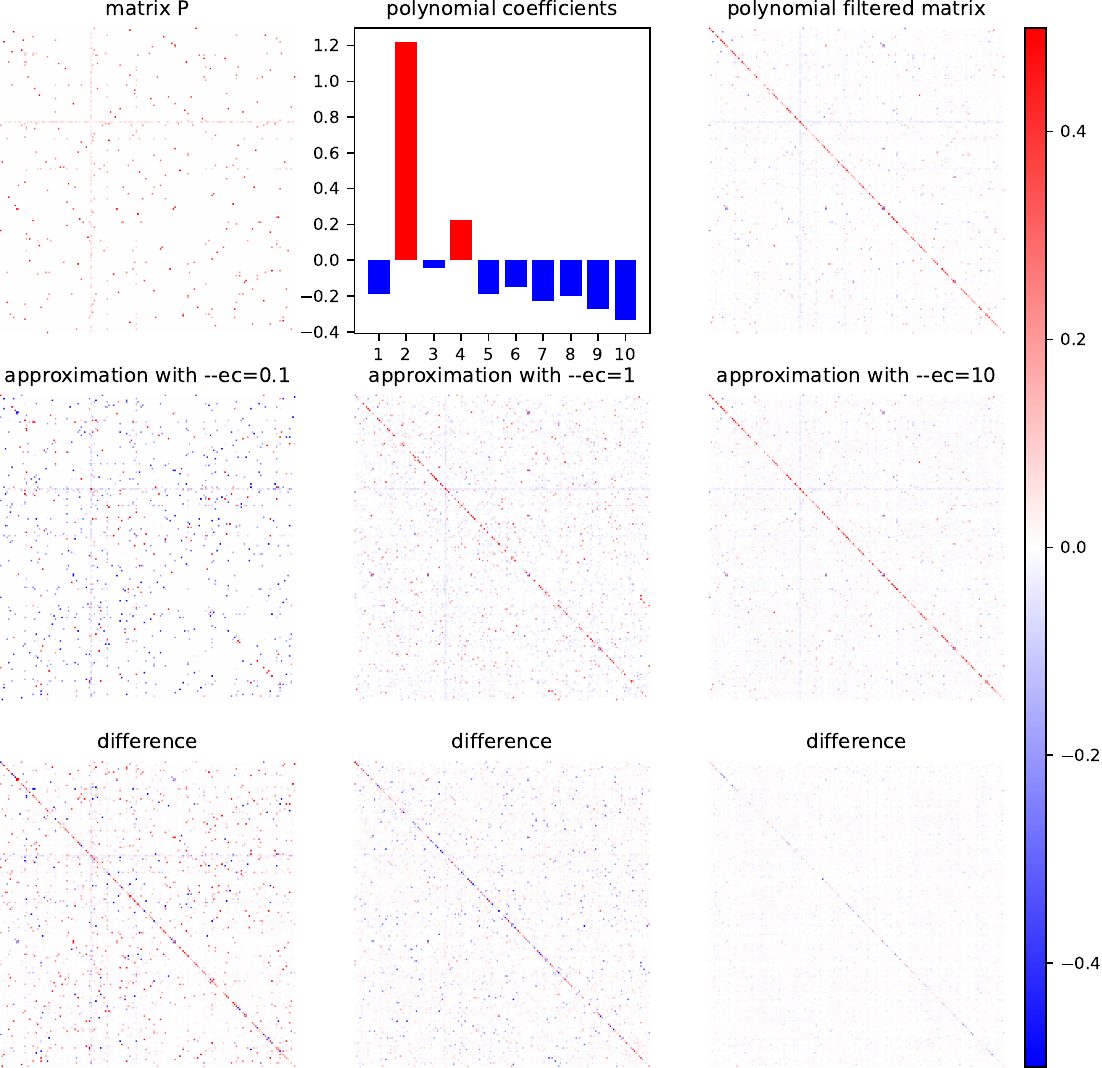}
  \caption{The visualization of the comparison between the polynomial filtered matrix and the results of the Laplacian sparsification with different numbers of samplings, on dataset Texas.}
  \label{fig:lap_similarity}
\end{figure}

In Figure~\ref{fig:lap_similarity}, we present a visualization to intuitively display the original matrix $\mathbf{P}$, the polynomial coefficients, and the difference between the polynomial of $\mathbf{P}$ and the approximated matrix with various sampling numbers on the Texas dataset. 
To mitigate the impact of dominant values, we excluded $\mathbf{P}^0$ (whose coefficient is 3.22) and fixed the range of the value bar to $[-0.5, 0.5]$.
As we have proved that our approximation is unbiased, the difference decreases with an increase in ``-\noindent-ec''.
When ``-\noindent-ec=10'', there is almost no difference between the approximated propagation matrix and the precisely calculated one, highlighting the promising results of our sampling method. The relationship between approximation similarity and final GNN performance is complex. 
More details about the sampling numbers can be found in Appendix~\ref{app:scale}.

\subsection{Results on Small-scale Real-world Datasets}
In this section, we conduct full-supervised transductive node classification tasks on 10 small-scale real-world datasets.
The detailed results are presented in the Table~\ref{tbl:res-small}.
Note that we evaluate the relative performance change between the original base model and its variant, with a clear distinction marked by the horizontal line.

% We first examine the comparison of GCN and SGC.
% SGC can be seen as the detached version of GCN, which first executes a $K$-hop graph propagation ($K=2$ in practice), followed by the linear layers.
% Our empirical findings verify that the detaching manner does exert a negative influence on the performance of GCN.
% Despite SGC exhibiting occasional performance improvements on specific small-scale datasets, the overall results are still distant from the powerful baseline GPR.

The entries in Table~\ref{tbl:res-small} offer an insightful comparison between some well-established GNN models, chosen spectral works, and their Laplacian sparsified variation.
GPR-GNN (abbreviated as GPR) is one of the strongest spectral GNNs with a learnable polynomial filter, making our evaluation promising.
The practical performance of GPR-GNN with Laplacian sparsification is superior to its original version, even under the potential risk of effect loss caused by the approximation.
We further investigate our method over non-trivial monomial bases and channel-wise filters, whose typical instances are JacobiConv and FavardGNN.
The results share a similar manner with the GPR-GNN ones.

We choose APPNP as the typical baseline for those spectral works with static polynomial filters.
In our implementation, the number of sampled edges for the Laplacian sparsification in spectral works with static polynomial filters is a mere fraction (specifically, $k$ times less) when compared to those with learnable filters.
Surprisingly, APPNP-LS exhibits equivalent or even superior performance compared to the original APPNP while achieving a similar performance elevation to the GPR series, despite employing fewer sampled edges.

GCN and SGC, serving as part of our baselines, provided us with some preliminary results about the comparison between common GNNs and their detached versions.
While SGC can be considered as the detached version of GCN, the detaching manner does negatively impact the performance of GCN.
A more detailed analysis of the detaching manner can be found in Appendix~\ref{app:detach}.

\begin{table}[t]
\caption{Experimental results of some baselines and our Laplacian sparsification entangled methods on \textbf{medium- and large-scale datasets}. Models with the suffix ``-de'' represent those with detached graph propagation and linear layer. Proposed results share the same annotation formats with Table~\ref{tbl:res-small}.}
\label{tbl:res-large}
\centering
\resizebox{0.9\linewidth}{!}{
\begin{tabular}{lcccc}
\toprule
Dataset & Arxiv & T-gamers & Penn94 & Papers100M\\
$\mathcal{H}(G)$ & 0.655 & 0.554 & 0.470 & -\\
\midrule
MLP         & 54.04 & 64.90 & \underline{83.17} & 41.36\\
GAT         & \underline{70.45} & 62.47 & 74.14 & GPU OOM\\
GCNII       & \underline{70.64} & 63.10 & 75.19 & GPU OOM\\
PPRGo       & 63.95 & 65.20 & \underline{84.46} & -\\
LMCGCN      & 67.52 & 61.61 & 81.57 & -\\
LazyGNN     & 70.26 & 60.58 & 74.49 & -\\
\midrule
GCN         & 70.18 & \textbf{67.86} & 82.55 & GPU OOM \\
SGC         & 70.35 & \underline{66.87} & GPU OOM & GPU OOM \\
SGC-de      & - & - & - & 59.26\\
$\Delta$    & +0.17 & -0.99 & - & - \\
\midrule
GPR         & \textbf{71.03} & \underline{66.69} & 82.92 & GPU OOM\\
GPR-de      & - & - & - & \underline{61.68}\\
GPR-LS      & 70.32 & 66.45 & \textbf{85.00} & \underline{61.76}\\
$\Delta$    & -0.71 & -0.24 & +2.08 & +0.08\\
\midrule
APPNP       & 69.87 & 65.11 & 82.89 & GPU OOM \\
APPNP-LS    & 69.08 & 65.83 & \underline{83.17} & \textbf{62.10}\\
$\Delta$    & -0.79 & +0.72 & +0.28 & - \\
\bottomrule
\end{tabular}
}
\vspace{-0.5em}
\end{table}

\subsection{Results on Large-scale Real-world Datasets}

To further assess the scalability of both our models and the baselines, we conduct experiments on a diverse range of medium- and large-scale real-world datasets.
Note that the dataset Penn94 contains high-dimensional original node representation, and Ogbn-papers100M includes an extensive network with over 111M nodes and 1.6B edges.
The experimental results are presented in Table~\ref{tbl:res-large}.

As is shown in Table~\ref{tbl:res-large}, our models consistently deliver competitive results even as we scale up to large datasets.
Interestingly, on the heterophilous datasets proposed by~\cite{DBLP:LINKX}, our models show a slight performance advantage over the corresponding base models, despite the inherent approximation imprecision.
This phenomenon shows that the Laplacian sparsification possesses the outstanding ability in 1) denoising, i.e. sampling the unnecessary neighbor connections with low probabilities, and 2) approximating the desired complex filters tailored to the heterophilous graphs.

As data scales up, many existing models suffer scalability challenges.
For instance, the standard SGC cannot execute forward propagation on Penn94 without preprocessing since Penn94 contains 1.4 million edges and 4,814 dimensions of original features.
The graph propagation leads to GPU out-of-memory errors without the dimensionality reduction.
The items marked with ``GPU OOM'' in Table~\ref{tbl:res-large} signify instances where the model cannot be trained on our devices.
For SGC and GPRGNN, we preprocess the graph propagation of the required hops on Ogbn-papers100M.
Our results clearly demonstrate that GPR-LS attains the performance of the corresponding decoupling method.
APPNP-LS allows APPNP to function normally within limited storage space.
These outcomes validate the effectiveness of our proposed method in significantly enhancing the scalability of conventional spectral approaches.

\begin{table}[t]
\caption{Experimental results of some baselines and our Laplacian sparsification entangled methods on dataset \textbf{MAG-scholar-C}. Proposed results share the same annotation formats with Table~\ref{tbl:res-small}.}
\label{tbl:res-mag}
\centering
\resizebox{\linewidth}{!}{
\begin{tabular}{lccccc}
\toprule
 & MAG-scholar-C  & Precomputation  & Average Epoch\\
 & Accuracy$\pm$std(\%)   & Time(s)        & Training Time(s)\\ 
\midrule
MLP         & 83.34$\pm$0.08 & 0 & 0.12\\
GCN         & GPU OOM & - & - \\
PPRGo       & \underline{87.15}$\pm$0.01 & 373.52 & 1.66\\
\midrule
SGC         & GPU OOM & - & -\\
SGC-de      & OOM & - & - \\
\midrule
GPR         & GPU OOM & - & -\\
GPR-de      & OOM & - & - \\
GPR-LS      & \textbf{87.30}$\pm$0.03 & 0 & 1.61\\
\midrule
APPNP       & GPU OOM & - & - \\
APPNP-LS    & \underline{86.67}$\pm$0.07 & 0 & 1.91\\
\bottomrule
\end{tabular}
}
\vspace{-0.5em}
\end{table}

\subsection{Results on MAG-scholar-C}
This section delves into the performance evaluation of the tested models on the MAG-scholar-C dataset, notable for its exceedingly high dimensionality of input node features (2.8M features). The outcomes of these experiments are summarized in Table~\ref{tbl:res-mag}.

It is evident from the experimental data that conventional GNN models are impractical for the MAG-scholar-C dataset due to the significant GPU memory constraints.
Moreover, detached models encounter limitations in propagating node features without prior dimensionality reduction, even when operated on CPUs equipped with 512GB of main memory.
Conversely, our proposed method enables the entangled models to process the MAG-scholar-C dataset efficiently using mini-batching and dimensionality reduction.

When compared to PPRGo, our GPR-LS model exhibits superior performance, achieving this with reduced training time.
Furthermore, all methods employing Laplacian sparsification in our study negate the necessity to compute the approximated PPR matrix, which is a computationally intensive task inherent in PPRGo.
Regarding the potential scalability issue possibly encountered when dealing with large-scale datasets, we provide an analysis about time, space, and excitability in Appendix~\ref{app:scale}

\section{Conclusion}
In this paper, we present a novel method centered on Laplacian sparsification to enhance the scalability of spectral GNNs significantly.
Our approach demonstrates its capability to approximate the equivalent propagation matrix of Laplacian filters, making it compatible with existing scalable techniques.
We provide the theoretical proof affirming that our model produces the correct sparsifier with probability at least $1-K/n$, approximation parameter $\varepsilon$, and $O\left(\frac{n\log n}{\varepsilon^2}\right)$ non-zeros in the propagation matrix, and thus conserves the core attributes of the propagated signal.
The experimental results validate that our methods yield comparable or even superior performance compared to the corresponding base models.
This remarkable achievement is particularly noteworthy considering that we sample far fewer edges than the theoretical bound, underscoring the exceptional ability of our method to approximate desired filters.

\bibliographystyle{ACM-Reference-Format}
\bibliography{sample-base}

%%
%% If your work has an appendix, this is the place to put it.
\appendix

\iffalse

\section{Research Methods}

\subsection{Part One}

Lorem ipsum dolor sit amet, consectetur adipiscing elit. Morbi
malesuada, quam in pulvinar varius, metus nunc fermentum urna, id
sollicitudin purus odio sit amet enim. Aliquam ullamcorper eu ipsum
vel mollis. Curabitur quis dictum nisl. Phasellus vel semper risus, et
lacinia dolor. Integer ultricies commodo sem nec semper.

\subsection{Part Two}

Etiam commodo feugiat nisl pulvinar pellentesque. Etiam auctor sodales
ligula, non varius nibh pulvinar semper. Suspendisse nec lectus non
ipsum convallis congue hendrerit vitae sapien. Donec at laoreet
eros. Vivamus non purus placerat, scelerisque diam eu, cursus
ante. Etiam aliquam tortor auctor efficitur mattis.

\section{Online Resources}

Nam id fermentum dui. Suspendisse sagittis tortor a nulla mollis, in
pulvinar ex pretium. Sed interdum orci quis metus euismod, et sagittis
enim maximus. Vestibulum gravida massa ut felis suscipit
congue. Quisque mattis elit a risus ultrices commodo venenatis eget
dui. Etiam sagittis eleifend elementum.

Nam interdum magna at lectus dignissim, ac dignissim lorem
rhoncus. Maecenas eu arcu ac neque placerat aliquam. Nunc pulvinar
massa et mattis lacinia.

\fi

\appendix
\section{Proofs of the Proposed Theories}
\label{app:theorem}
\subsection{Detailed Analysis about SLSGC and GLSGC Algorithm}

In this section, we begin by restating the previously established conclusion.

\begin{theorem}
    \label{the:RWspar}
    ~\citep{DBLP:sparsifierRWP} Given a weighted graph $G$ and the upper bound of its effective resistance $R_{\text{sup}}(e)\ge R(e)$. For any approximation parameter $\varepsilon$, there exists a sampled graph $\widetilde{G}$ with at most $M=O\left(\log n/\varepsilon^2 \cdot \left(\sum_{e\in E}w(e)R_{\text{sup}}(e)\right)\right)$ edges, satisfying $(1-\varepsilon)\mathbf{L}_G\preccurlyeq \mathbf{L}_{\widetilde{G}} \preccurlyeq (1+\varepsilon)\mathbf{L}_G$ with at least $1-1/n$ probability.
\end{theorem}

This theorem marks the initial step, whose pseudo-code of the algorithm is presented at Algorithm~\ref{alg:edgesamp_ER}.
Even though our focus is on an unweighted graph $G$, the final destination of our approximation $\mathbf{D}(\mathbf{D}^{-1}\mathbf{A})$ is inherently weighted.
The weight between node $u$ and $v$ on the graph $G_{r}$ with its adjacency matrix $\mathbf{A}_{r}=\mathbf{D}(\mathbf{D}^{-1}\mathbf{A})^r$ can be considered as the union of all the paths between nodes $u$ and $v$ with length $r$ showing up in the unweighted graph $G$, i.e. $\mathbf{A}_r(u,v)=\sum_{p\in P} w(p)$, where 
\begin{equation*}
    P=\left\{(u=i_0, i_1, \cdots, i_{r-1}, v=i_r) | (i_j, i_{j+1})\in E, j=0,1,\cdots, r-1\right\}.
\end{equation*}

The weight of a path, denoted as $p=(u_0, u_1,..., u_r)$ can be formally expressed as
\begin{equation*}
    w(p)=\frac{\prod_{i=0}^{r-1} \mathbf{A}_{u_{i-1},u_i}}{\prod_{i=1}^{r-1}\mathbf{D}_{u_i,u_i}}.
\end{equation*}
This value is symmetrical when viewed from $u_0$ and $u_r$, which is a slight deviation from the random walk probability, as it encompasses the probability of the walk starting from the initial node.

Retrieving Lemma~\ref{lem:ERupperbound}, the upper bound of effective resistance for a path $p$ on the weighted graph can be expressed as
\begin{equation*}
    R_{\text{sup},G_r}(u_0,u_r) = \sum_{i=0}^{r-1} \frac{2}{\mathbf{A}_{u_i,u_{i+1}}},
\end{equation*}
where $p$ is a path within graph $G$, startingfrom $u_0$, ending at $u_r$, and possessing a length of $r$.

We derive the conclusion that 
\begin{align}
    &\sum_{p=(u_0,\cdots,u_r)} w(p)R_{\text{sup},G_r}(u_0,u_r) \\
    &= \sum_p \left(\sum_{i=0}^{r-1}\frac{2}{\mathbf{A}_{u_i,u_{i+1}}}\right)\left(\frac{\prod_{j=0}^{r-1}\mathbf{A}_{u_j,u_{j+1}}}{\prod_{j=1}^{r-1}\mathbf{D}_{u_j,u_j}}\right)\notag\\
    &= 2 \sum_p \sum_{i=1}^r \left(\frac{\prod_{j=1}^{i-1}\mathbf{A}_{u_{j-1}, u_j}\prod_{j=i}^{r-1}\mathbf{A}_{u_j, u_{j+1}}}{\prod_{j=1}^{r-1}\mathbf{D}_{u_j,u_j}}\right)\notag\\
    &= 2 \sum_{e\in E} \sum_{i=1}^r\left(\sum_{p|(u_{i-1},u_i)=e} \frac{\prod_{j=1}^{i-1}\mathbf{A}_{u_{j-1,u_j}}}{\prod_{i=1}^{j-1}\mathbf{D}_{u_j,u_j}}\cdot\frac{\prod_{j=i}^{r-1}\mathbf{A}_{u_{j,u_{j+1}}}}{\prod_{j=i}^{r-1}\mathbf{D}_{u_j,u_j}} \right) \label{eqn:rwscenter}\\
    &= 2\sum_{e\in E}\sum_{i=1}^r 1\notag\\
    &= 2mr \label{eqn:final2mr}.
\end{align}
This conclusion holds since $\mathbf{D}_{u,u} = \sum_{v\in V} \mathbf{A}_{u,v}$.
Recall that the derivation of Equation~\ref{eqn:rwscenter} implies the sampling method process of the Laplacian sparsification.

For the proposed Algorithm~\ref{alg:edgesamp}, the probability of sampling a distinct path $p=(u_0,\cdots,u_r)$ can be derived as

\begin{align*}
    &\text{Pr}(p=(u_0,\cdots,u_r|e,k))\\
    &=\text{Pr}((u_{k-1},u_k)=e)\cdot \text{Pr}(p=(u_0,\cdots,u_r)|(u_{k-1},u_{k})=e)\\
    &= \frac{1}{m}\cdot \prod_{i=1}^{k-1}\frac{\mathbf{A}_{u_i,u_{i-1}}}{\mathbf{D}_{u_i,u_i}}\cdot \prod_{i=k}^{r-1} \frac{\mathbf{A_{u_i, u_{i+1}}}}{\mathbf{D}_{u_i,u_i}}\\
    &= \frac{1}{m} \cdot \frac{\prod_{i=1}^r \mathbf{A}_{u_{i-1},u_i}}{\left(\prod_{i=1}^{r-1}\mathbf{D}_{u_i,u_i}\right)\cdot \mathbf{A}_{u_{k-1},u_k}}.
\end{align*}

Since $e$ is sampled uniformly at random, as indicated by the term $\text{Pr}((u_{k-1},u_k)=e)$ above, we now consider the randomness introduced by $k$.
$k$ is also sampled uniformly at random. Thus, eliminating $k$ finally yields

\begin{align*}
    &\text{Pr}(p=(u_0,\cdots,u_r))\\
    &=\left(\frac{1}{m}\cdot\frac{\prod_{i=1}^{r}\mathbf{A}_{u_{i-1},u_i}}{\prod_{i=1}^{r-1}\mathbf{D}_{u_i,u_i}}\right)\left(\frac{1}{r}\cdot\sum_{k=1}^r\frac{1}{\mathbf{A}_{u_{k-1},u_k}}\right)\\
    &=\frac{1}{2mr}w(p)R_{\text{sup},G_r}(p).
\end{align*}

Since all the edge weights are considered as $1$ on unweighted graphs, the upper bound $R_{\text{sup},G_r}$ is $2r$ for any $p$ with length $r$.
As is proved that $\text{Pr}(p)\varpropto w(e)R_{\text{sup}}(p)$, we can execute the Algorithm~\ref{alg:edgesamp_ER} with $M$ times of Monte-Carlo sampling stated in Algorithm~\ref{alg:edgesamp} to construct an $\varepsilon$-sparsifier.

In another view, the probability of sampling a distinct path $p$ is proportional to $w(p)$.
Follow Equation~\ref{eqn:final2mr}, the summation of $w(p)$ for any length $r$ is $m$.
We can execute the Monte-Carlo sampling $M$ times to generate $M$ paths with length $r$ and weight $m/M$ to generate the unbiased approximation of $\mathbf{D}(\mathbf{D}^{-1}\mathbf{A})^r$, which further proves the accuracy of the sampling procedure. 

Hence, Theorem~\ref{the:RWspar} have been proven. 
There exists an algorithm that employs the process outlined above to generate an $\varepsilon$-sparsifier of $\mathbf{D}(\mathbf{D}^{-1}\mathbf{A})^r$ within $M=O(n\log n/\varepsilon^2)$ edges with the probability $1-1/n$ at least.
Our final objective is to approximate the $\mathbf{D}^{-1/2}\cdot\mathbf{D}\left(\sum_{k=0}^K w_k\left(\mathbf{D}^{-1}\mathbf{A}\right)^k \right)\mathbf{D}^{-1/2}$ proposed in Equation~\ref{Eqn:SpecFilter}.

In general, all of the sampled edges $(u_0, u_r)$, representing path $p=(u_0,...,u_r)$, should be multiplied by a correction coefficient of $d_{u_0}^{-1/2}d_{u_r}^{-1/2}$, intuitively. 

For random walk polynomials with static coefficients, such as in SLSGC, one effective method to approximate the middle term is to distribute all the edges with the probability proportional to $|w_i|$ for each path length $i$.
Hence, we can first pick the length $i$ with the probability $\text{Pr}\{r=i\}=|w_i| / \Vert\mathbf{w}\Vert_1$, then sample the edge and correct the edge weight with $d_{u_0}^{-1/2} d_{v_0}^{-1/2} \text{sgn}(w_i)\Vert\mathbf{w}\Vert_1$ to compensate for the rescaling of the probability.
This process intuitively maintains the unbiased approximation and the Laplacian sparsification properties.
In conclusion, when we are approximating the graph $\mathbf{D}(\mathbf{D}^{-1}\mathbf{A})^k$, the edge value is adjusted to
\begin{equation*}
    \text{sgn}(w_k)\Vert\mathbf{w}\Vert_1\cdot d_u^{-1/2}d_v^{-1/2}\cdot m/M.
\end{equation*}

For random walk polynomials with learnable coefficients, the models are required to generate the correct hop-independent derivative of the $w_i$.
This limitation prevents us from directly sampling with $w_k$.
Instead, we may sample the graph hop-by-hop, meaning that for each random walk length $i$, we independently generate the sparsifier of graph $\mathbf{D}(\mathbf{D}^{-1}\mathbf{A})^i$ and stack them with corresponding coefficients $w_i$.

To maintain the property of being an $\varepsilon$-sparsifier of the given graph, one sufficient condition is that all the generated $K$ sparsifiers are $\varepsilon$-sparsifiers.
Thus, the required number of generated edges increases to 
\begin{equation*}
    K\cdot O(n\log n/\varepsilon^2)=O(n\log n/\varepsilon^2),
\end{equation*}
and the probability decreases to 
\begin{equation*}
    (1-1/n)^K\ge 1-K/n.
\end{equation*}
Since $K$ is a predefined constant and $K\ll n$, the complexity does not change significantly.
Meanwhile, the generated $K$ components are all unbiased approximations of $\mathbf{D}(\mathbf{D}^{-1}A)^i$, the stacked approximation is also unbiased, evidently.
Thus, Theorem~\ref{theorem:rwps} is proved.

\subsection{Proof of Theorem~\ref{the:nodewise}}

We can begin by considering the probability of the sampling method selecting a distinct path on the graph.
Assume the desired path is $p=(u_0,\cdots,u_r)$ and the first selected node is $u$, we can derive the probability as follows:
\begin{align*}
    &\text{Pr}(p=(u_0,\cdots,u_r)) \\
    &= \text{Pr}(u=u_0) \cdots \text{Pr}(p=(u_0,\cdots,u_r)|u=u_0)\\
    &= \frac{\mathbf{D}_{u_0,u_0}}{\sum_{v\in U}\mathbf{D}_{v,v}} \cdot \prod_{i=0}^{r-1}\frac{\mathbf{A}_{u_i,u_{i+1}}}{\mathbf{D}_{u_i,u_i}}\\
    &= \frac{1}{\sum_{v\in U}\mathbf{D}_{v,v}} \cdot\frac{\prod_{i=0}^{r-1}\mathbf{A}_{u_i,u_{i+1}}}{\prod_{i=1}^{r-1}\mathbf{D}_{u_i,u_i}}\\
    &= \frac{w(p)}{\sum_{v\in U}\mathbf{D}_{v,v}},
\end{align*}
where $U$ is the set of nodes where the start of random walks are selected from. 

From the derivation of Equation~\ref{eqn:final2mr}, we can conclude that the summation of weights of all the paths starting from node $u$ can be divided into $\mathbf{D}_{u,u}$ series, where each series starts with one of the $\mathbf{D}_{u,u}$ edges incident with node $u$.
Since the summation of all the paths with one distinct $k$ satisfying $(u_k, u_{k+1})=e$ is $1$, we have the summation of weights of all the paths starting from $u$ is $\mathbf{D}_{u,u}$.
This means the probability of sampling a distinct path is proportional to its weight.

The entry $\left(\mathbf{D}(\mathbf{D}^{-1}\mathbf{A})^r\right)_{i,j}$ represents the combination of the weights of all the path $p=(u_0=i,u_1,\cdots,u_{r-1},u_r=j)$.
For each sampled path, we can correct it by multiplying it with $\frac{1}{M}\sum_{v\in U}\mathbf{D}_{v,v}$ to obtain an unbiased approximation of the weight of each path.
The union of the paths will inevitably generate an unbiased approximation of the corresponding rows of $U$ in $\mathbf{D}(\mathbf{D}^{-1}A)^r$.
Thus, the final weight generated for the path $p=(u_0, \cdots,u_r)$ is 
\begin{equation*}
    \frac{d_{u_0}^{-1/2}d_{u_r}^{-1/2}}{M}\left(\sum_{v\in U}\mathbf{D}_{v,v}\right),
\end{equation*}
which indicates that the Theorem~\ref{the:nodewise} has been proven.

\subsection{Proof of Theorem~\ref{the:appnplosserror}}

First, the loss function can be derived as
\begin{align*}
    \mathcal{L}(\mathbf{z})=(1-\alpha)\mathbf{z}^\top\mathbf{L}\mathbf{z}+\alpha(\mathbf{z-x})^\top(\mathbf{z-x}).
\end{align*}

Let $f(\mathbf{P}, K)=\sum_{k=0}^{K-1} \alpha(1-\alpha)^k\mathbf{P}^k+(1-\alpha)^K\mathbf{P}^K$ be the original APPNP propagation matrix, and $\tilde{f}(\mathbf{P}, K)$ be the approximated one. 
Evidently, the connection between the normalized Laplacian and the propagation matrix can be expressed as:
\begin{align*}
    \mathbf{L}_{K} = \mathbf{I} - f(\mathbf{P}, K),\\
    \widetilde{\mathbf{L}}_{K} = \mathbf{I} - \tilde{f}(\mathbf{P}, K).
\end{align*}

For convenience, we abbreviate $f(\mathbf{P}, K)$ and $\tilde{f}(\mathbf{P}, K)$ into $f$ and $\tilde{f}$, respectively.
The propagated signals are defined as $\mathbf{z}=f\mathbf{x}$ and $\tilde{\mathbf{z}}=\tilde{f}\mathbf{x}$.
Then, loss functions $\mathcal{L}(\mathbf{z})$ and $\mathcal{L}(\tilde{\mathbf{z}})$ can be derived as:
\begin{align*}
    \mathcal{L}(\mathbf{z}) 
    &= (1-\alpha)\mathbf{x}^\top f^\top \mathbf{L} f\mathbf{x} + \alpha\mathbf{x}^\top(f-\mathbf{I})^\top(f-\mathbf{I})\mathbf{x}\\
    &= \mathbf{x}^\top\left((1-\alpha) f^\top \mathbf{L} f + \alpha(f-\mathbf{I})^2\right)\mathbf{x},\\
    \mathcal{L}(\tilde{\mathbf{z}})
    &= \mathbf{x}^\top\left((1-\alpha) \tilde{f}^\top \mathbf{L} \tilde{f} + \alpha(\tilde{f}-\mathbf{I})^2\right)\mathbf{x}.
\end{align*}

According to the conditions, we can obtain some partial orderings between $f$ and $\tilde{f}$:
\begin{align*}
    \begin{cases}
        (1-\varepsilon)\left(\mathbf{I}-f\right) \preccurlyeq \mathbf{I}-\tilde{f} \preccurlyeq (1+\varepsilon)\left(\mathbf{I}-f\right) \\
        (1-\varepsilon)^2\left(\mathbf{I}-f\right)^2 \preccurlyeq (\mathbf{I}-\tilde{f})^2 \preccurlyeq (1+\varepsilon)^2\left(\mathbf{I}-f\right)^2 \\
        -\varepsilon(\mathbf{I}-f)\preccurlyeq\Delta\preccurlyeq\varepsilon(\mathbf{I}-f)\\
        -\frac{\varepsilon}{1+\varepsilon}(\mathbf{I}-\tilde{f}) \preccurlyeq \Delta \preccurlyeq \frac{\varepsilon}{1-\varepsilon}(\mathbf{I}-\tilde{f})
    \end{cases},
\end{align*}
where $\Delta=\mathbf{L}_K-\widetilde{\mathbf{L}}_{K}=f-\tilde f$.
Thus, we may calculate the absolute error between $\mathcal{L}(\mathbf{z})$ and $\mathcal{L}(\mathbf{\tilde{z}})$ as follows:
\begin{align*}
    \mathcal{L}(\mathbf{z})-\mathcal{L}(\mathbf{\tilde{z}})
    &=\mathbf{x}^\top\left((1-\alpha) \left(f^\top \mathbf{L} f-\tilde f^\top \mathbf{L} \tilde f\right) + \alpha\left((f-\mathbf{I})^2-(\tilde{f}-\mathbf{I})^2\right)\right)\mathbf{x}\\
    &= \mathbf{x}^\top\left((1-\alpha)\delta\mathbf{X}+\alpha \delta\mathbf{Y}\right)\mathbf{x},\\
    \delta\mathbf{X}&=f^\top \mathbf{L} f-\tilde f^\top \mathbf{L} \tilde f\\
    &= f \mathbf{L} f - f\mathbf{L}\tilde f + f\mathbf{L}\tilde f - \tilde f \mathbf{L} \tilde f\\
    &= f \mathbf{L} (f-\tilde f) + (f-\tilde f)\mathbf{L}\tilde f\\
    &= f\mathbf{L}\Delta + \Delta\mathbf{L}\tilde{f},\\
    \delta\mathbf{Y} &= (\mathbf{I}-f)^2 - (\mathbf{I}-\tilde{f})^2.
\end{align*}

Then we calculate the relative error $\left|\frac{\mathcal{L}(z)-\mathbf{\mathcal{L}(\tilde{z})}}{\mathcal{L}(z)}\right|$ by part:
\begin{align*}
    \frac{\mathbf{x}^\top\delta\mathbf{X}\mathbf{x}}{\mathbf{x}^\top f^\top \mathbf{L} f\mathbf{x}} 
    &=  \frac{\mathbf{x}^\top(f\mathbf{L}\Delta + \Delta L\tilde{f})\mathbf{x}}{\mathbf{x}^\top f^\top \mathbf{L} f\mathbf{x}}.
\end{align*}

Since the matrices $\Delta, \mathbf{L}, \tilde{f}$ are all symmetric, we have $\mathbf{x}^\top\Delta\mathbf{L}\tilde{f}\mathbf{x}=\mathbf{x}^\top(\Delta\mathbf{L}\tilde{f})^\top\mathbf{x}=\mathbf{x}^\top\tilde{f}\mathbf{L}\Delta\mathbf{x}$.
Thus, we can derive that:
\begin{align*}
    \frac{\mathbf{x}^\top\delta\mathbf{X}\mathbf{x}}{\mathbf{x}^\top f^\top \mathbf{L} f\mathbf{x}} 
    &=  \frac{\mathbf{x}^\top(f+\tilde{f})\mathbf{L}\Delta\mathbf{x}}{\mathbf{x}^\top f^\top \mathbf{L} f\mathbf{x}}\\
    &= \frac{2\mathbf{x}^\top f\mathbf{L}\Delta\mathbf{x}}{\mathbf{x}^\top f^\top \mathbf{L} f\mathbf{x}} - \frac{\mathbf{x}\Delta\mathbf{L}\Delta\mathbf{x}^\top}{\mathbf{x}^\top f^\top \mathbf{L} f\mathbf{x}}\\
    &= O(\varepsilon) - O(\varepsilon^2) = O(\varepsilon).
\end{align*}

Meanwhile, the other part shows
\begin{align*}
    \frac{\mathbf{x}^\top\delta Y\mathbf{x}}{\mathbf{x}^\top(\mathbf{I}-f)^2\mathbf{x}}
    &=\frac{\mathbf{x}^\top\left((\mathbf{I}-f)^2-(\mathbf{I}-\tilde{f})^2\right)\mathbf{x}}{\mathbf{x}^\top(\mathbf{I}-f)^2\mathbf{x}}\\
    &=1-\frac{\mathbf{x}^\top(\mathbf{I}-\tilde{f})^2\mathbf{x}}{\mathbf{x}^\top(\mathbf{I}-f)^2\mathbf{x}}\\.
\end{align*}

Since we have
\begin{align*}
&(1-\varepsilon)^2\left(\mathbf{I}-f\right)^2 \preccurlyeq (\mathbf{I}-\tilde{f})^2 \preccurlyeq (1+\varepsilon)^2\left(\mathbf{I}-f\right)^2,
\end{align*}
we may obtain
\begin{align*}
\Rightarrow&-2\varepsilon-\varepsilon^2 \le 1-\frac{\mathbf{x}^\top(\mathbf{I}-\tilde{f})^2\mathbf{x}}{\mathbf{x}^\top(\mathbf{I}-f)^2\mathbf{x}} \le 2\varepsilon-\varepsilon^2,\\
\Rightarrow&1-\frac{\mathbf{x}^\top(\mathbf{I}-\tilde{f})^2\mathbf{x}}{\mathbf{x}^\top(\mathbf{I}-f)^2\mathbf{x}} = O(\varepsilon)+O(\varepsilon^2) = O(\varepsilon).
\end{align*}

Hence, the relative error finally yields
\begin{align*}
    \left|\frac{\mathcal{L}(z)-\mathbf{\mathcal{L}(\tilde{z})}}{\mathcal{L}(z)}\right| 
    &= \left|\frac{\mathbf{x}^\top\left((1-\alpha)\delta\mathbf{X}+\alpha \delta\mathbf{Y}\right)\mathbf{x}}{\mathbf{x}^\top\left((1-\alpha) f^\top \mathbf{L} f + \alpha(f-\mathbf{I})^2\right)\mathbf{x}}\right|\\
    &\le \left|\frac{\mathbf{x}^\top(1-\alpha)\delta\mathbf{X}\mathbf{x}}{\mathbf{x}^\top\left((1-\alpha) f^\top \mathbf{L} f + \alpha(f-\mathbf{I})^2\right)\mathbf{x}}\right| \\
    &+ \left|\frac{\mathbf{x}^\top \alpha\delta\mathbf{Y}\mathbf{x}}{\mathbf{x}^\top\left((1-\alpha) f^\top \mathbf{L} f + \alpha(f-\mathbf{I})^2\right)\mathbf{x}}\right|\\
    &\le \left|\frac{\mathbf{x}^\top(1-\alpha)\delta\mathbf{X}\mathbf{x}}{\mathbf{x}^\top(1-\alpha) f^\top \mathbf{L} f\mathbf{x}}\right|
    + \left|\frac{\mathbf{x}^\top \alpha\delta\mathbf{Y}\mathbf{x}}{\mathbf{x}^\top\alpha(f-\mathbf{I})^2\mathbf{x}}\right| \\
    &= O(\varepsilon) + O(\varepsilon) = O(\varepsilon).
\end{align*}

\subsection{Time and Space Complexity Analysis}
\label{app:overheadanalysis}

In this section, we provide a thorough time and space complexity analysis of our methods and the used backbones. 

Briefly, our method streamlines the graph convolution process in spectral GNNs. For $k$-order polynomial filters, the conventional SGNNs require $O(Km)$ node message-passing operations, whereas our approach reduces this to $O(Kn\log n/\varepsilon^2)$. GPR-GNN serves as our primary model for detailed analysis, with implications extendable to similar models like FavardGNN.

\header{\bf Time and memory overhead of GPRGNN.}

Training time: $O(KmF+LnF^2+nF_iF)$ per epoch, where $F_i$ denotes the dimension of original node embeddings.

Training memory: $O(LnF+LF^2+F_iF+m)$.

\header{\bf For detached GPRGNN, the precomputation of propagations are required.}

Precomputation time: $O(KmF_i)$ for $K$ rounds of graph propagation.

Precomputation memory: $O(KnF_i)$ to store $K+1$ node representation matrices of different hops.

Training time: $O(Ln_tF^2+n_tF_iF)$ per \textbf{epoch}, where $n_t$ is the number of nodes in the training set. The training contains $L$ layers of MLP.

Training memory: $O(Ln_bF+LF^2+F_iF)$ per \textbf{batch}. This phase includes $L$ Layers of MLP, while the initial layer transform $F_i$ dimensional layers to $F$ dimensions. $n_b$ denotes the number of nodes in mini-batch, which implies that the model can be trained with mini-batch.

\header{\bf For GPRGNN-LS, the whole model is entangled.}

Training time: $O(KMF+LnF^2+F_sF)$ per \textbf{epoch}, where $M=O(n\log n/\varepsilon^2)$, and $F_s$ determines the number of non-zeros in all of the original node embeddings. 

Sampling time: $O(K^2M)=O(K^2n\log n/\varepsilon^2)$, which is not the bottleneck since the random walk number can be controlled without influencing the performance negatively.

Training time: $O(Kn_tr_sF+LnF^2+F_{st}F)$ per \textbf{epoch} for semi-supervised tasks, where $r_s$ is the number of random walk sampling, and $F_{st}$ is the number of non-zeros in all of the training node original embeddings.

Sampling time: $O(K^2n_tr_s)$.

Training memory: $O(Ln_bF+LF^2+F_iF+m)$ per \textbf{batch}.

Our proposed method addresses the significant GPU memory constraints encountered with very large graphs, such as the Ogbn-papers100M dataset, where storing node embeddings for the entire graph requires $O(LnF)$ memory. 
By enabling mini-batch training for SGNNs, we effectively mitigate these limitations.
As our comprehensive time complexity analysis employs GPR-GNN as the primary example, the conclusions drawn from this model are applicable to others, such as FavardGNN, by similar inference.

For datasets with dense original node embeddings, such as $F_s \approx nF_i$ and $F_{st} \approx n_tF_i$, our method remains efficient. However, in cases like the MAG-Scholar-C dataset, where $F_s \ll nF_i$ and $F_{st} \ll n_tF_i$, traditional approaches like detached GPRGNN encounter bottlenecks due to the high computational demands of graph propagation on CPUs, even with substantial RAM. Our approach overcomes these challenges by eliminating the need for extensive precomputation in the detached GPRGNN and facilitating mini-batch training, thus accommodating large-dimensional sparse node embeddings more effectively.

\section{Example Pseudo-codes of the Entangled Models}

\begin{algorithm}[t]
    \DontPrintSemicolon
    \caption{{\sf Static Laplacian Sparsified Graph Construction (SLSGC)}} \label{alg:SLSGC}
    \KwIn{Hop weights $\mathbf{w}$.}
    \KwIn{{\bf Model Saved:} Vertice set $V$, edge set $E$, degrees $\mathbf{d}$, maximum neighbor hop $K$, total sampling number $M$.}
    %\KwIn{\textbf{Model Saved: }Graph $G$, Degree $\mathbf{d}$, Historical Embeddings $\overline{\mathbf{X}}$, Neighbor Hop $K$, Hop Weight $\mathbf{w}$, Number of Walks $r_s$.}
    \KwOut{Weighted edge set $\widetilde{E}$ after laplacian sparsification.}
    $m\gets |E|$\;
    $\widetilde{E} \gets \emptyset$ \;
    \For {$i$ {\rm from} $1$ {\rm to} $M$}{
        $k\gets$ sample an integer $k$ from distribution $\text{Pr}\{k=j\}=\vert w_j\vert / \Vert \mathbf{w}\Vert_1$\;
        $(u,v)\gets$ Edge\_Sampling($E,k$)\;
        $\widetilde{E} \gets \widetilde{E}\cup \left((u, v), \text{sgn}(w_k)\Vert \mathbf{w}\Vert_1d_u^{-1/2}d_v^{-1/2}\cdot \frac{m}{M}\right)$ \;
    }
    \Return $\widetilde{E}$\;
    
\end{algorithm}

\begin{algorithm}[t]
    \DontPrintSemicolon
    \caption{{\sf APPNP with Laplacian Sparsification}} \label{alg:APPNP-LS}
    \KwIn{Node embeddings $\mathbf{X}$, training status $T$.}
    \KwIn{{\bf Model Saved:} Vertice set $V$, edge set $E$, degrees $\mathbf{d}$, maximum neighbor hop $K$, sampling number $M$, hop weight matrix $\mathbf{W}$.}
    \KwOut{Processed node embeddings $\widetilde{\mathbf{X}}$ }
    $\mathbf{X}\gets \text{Linear}(\mathbf{X})$\;
    \If {T}{
        $\widetilde{E}\gets\text{SLSGC}(\mathbf{W}_0)$\;
        $\widetilde{\mathbf{X}}\gets$ A round of message passing of $\mathbf{X}$ on edge set $\widetilde{E}$\;
    }
    \Else{
        $\widetilde{\mathbf{X}}\gets \mathbf{W}_{0,0}\mathbf{X}$\;
        \For {$i$ {\rm from} $1$ {\rm to} $K$}{
            $\mathbf{X}\gets$ A round of message passing of $\mathbf{X}$ on edge set $G.E$ with weights $\mathbf{D}^{-1/2}\mathbf{A}\mathbf{D}^{-1/2}$\;
            $\widetilde{\mathbf{X}}\gets \widetilde{\mathbf{X}}+\mathbf{W}_{0,i}\mathbf{X}$
        }
    }
    \# Possibly move ahead.\;
    $\widetilde{\mathbf{X}}\gets \text{Linear}(\widetilde{\mathbf{X}})$\; 
    \Return $\widetilde{\mathbf{X}}$
\end{algorithm}

\subsection{SLSGC and APPNP}
\label{app:APPNP-LS}
This section presents the pseudo-code of constructing a Laplacian spasified random walk polynomial with static coefficients in Algorithm~\ref{alg:SLSGC} and Laplacian sparsification entangled APPNP in Algorithm~\ref{alg:APPNP-LS}.
Note that the hop weight matrix $\mathbf{W}$ is not learnable.
$\mathbf{W}$ is predefined as $\mathbf{W}_{0,i}=\alpha(1-\alpha)^{i},i\ne K$ and $\mathbf{W}_{0,K}=(1-\alpha)^K$, where $\alpha$ is a hyper-parameter.

\begin{algorithm}[t]
    \DontPrintSemicolon
    \caption{{\sf General Laplacian Sparsified Graph Construction (GLSGC)}} \label{alg:GLSGC}
    \KwIn{Hop weights $\mathbf{w}$.}
    \KwIn{{\bf Model Saved:} Vertice set $V$, edge set $E$, degrees $\mathbf{d}$, maximum neighbor hop $K$, layer-wise sampling number $M$.}
    %\KwIn{\textbf{Model Saved: }Graph $G$, Degree $\mathbf{d}$, Historical Embeddings $\overline{\mathbf{X}}$, Neighbor Hop $K$, Hop Weight $\mathbf{w}$, Number of Walks $r_s$.}
    \KwOut{Weighted edge set $\widetilde{E}$ after laplacian sparsification.}
    $m, \widetilde{E}\gets |E|, \emptyset$\;
    \For{{\rm each} $v\in V$}{
        $\widetilde{E} \gets \widetilde{E}\cup ((v, v), \mathbf{w}_0)$ \;
    }
    \For {$k$ {\rm from} $1$ {\rm to} $K$}{
        \For {$j$ {\rm from} $1$ {\rm to} $M$}{
            $(u,v)\gets$ Edge\_Sampling($E$, $k$)\;
            $\widetilde{E} \gets \widetilde{E}\cup \left((u, v), \mathbf{w}_kd_u^{-1/2}d_v^{-1/2}\cdot \frac{m}{M}\right)$ \;
        }
    }
    \Return $\widetilde{E}$\;
\end{algorithm}

\begin{algorithm}[t]
    \DontPrintSemicolon
    \caption{{\sf GPRGNN with Laplacian Sparsification}} \label{alg:GPR-LS}
    \KwIn{Node embeddings $\mathbf{X}$, training status $T$.}
    \KwIn{{\bf Model Saved:} Vertice set $V$, edge set $E$, degrees $\mathbf{d}$, maximum neighbor hop $K$, sampling number $M$, hop weight matrix $\mathbf{W}$.}
    \KwOut{Processed node embeddings $\widetilde{\mathbf{X}}$ }
    $\mathbf{X}\gets \text{Linear}(\mathbf{X})$\;
    \If {T}{
        $\widetilde{E}\gets\text{GLSGC}(\mathbf{W}_0)$\;
        $\widetilde{\mathbf{X}}\gets$ A round of message passing of $\mathbf{X}$ on edge set $\widetilde{E}$\;
    }
    \Else{
        $\widetilde{\mathbf{X}}\gets \mathbf{W}_{0,0}\mathbf{X}$\;
        \For {$i$ {\rm from} $1$ {\rm to} $K$}{
            $\mathbf{X}\gets$ A round of message passing of $\mathbf{X}$ on edge set $G.E$ with weights $\mathbf{D}^{-1/2}\mathbf{A}\mathbf{D}^{-1/2}$\;
            $\widetilde{\mathbf{X}}\gets \widetilde{\mathbf{X}}+\mathbf{W}_{0,i}\mathbf{X}$
        }
    }
    \# Possibly move ahead\;
    $\widetilde{\mathbf{X}}\gets \text{Linear}(\widetilde{\mathbf{X}})$\;
    \Return $\widetilde{\mathbf{X}}$
\end{algorithm}

\subsection{GLSGC and GPRGNN-LS}
\label{app:GPR-LS}
This section presents the pseudo-code of constructing a Laplacian sparsified random walk matrix polynomial with learnable coefficients in Algorithm~\ref{alg:GLSGC} and Laplacian sparsification entangled GPRGNN in Algorithm~\ref{alg:GPR-LS}.
Note that the hop weight matrix $\mathbf{W}$ is learnable.
Please follow GPRGNN~\citep{DBLP:GPR-GNN} for more details of the initialization of $\mathbf{W}$.

\begin{algorithm}[t]
    \DontPrintSemicolon
    \caption{{\sf Node-Wise Laplacian Sampling}} \label{alg:semi}
    \KwIn{Hop weights $\mathbf{w}$, batch $B$}
    \KwIn{{\bf Model Saved:} Vertice set $V$, edge set $E$, degrees $\mathbf{d}$, maximum neighbor hop $K$, sampling number $M$.}
    %\KwIn{\textbf{Model Saved: }Graph $G$, Degree $\mathbf{d}$, Historical Embeddings $\overline{\mathbf{X}}$, Neighbor Hop $K$, Hop Weight $\mathbf{w}$, Number of Walks $r_s$.}
    \KwOut{Weighted edge set $\widetilde{E}$ after random walk sampling.}
    $s\gets \sum_{u\in B} \mathbf{d}_u$\;
    $\widetilde{E}\gets \emptyset$\;
    \For {$k$ {\rm from} $1$ {\rm to} $K$}{
        \For {$j$ {\rm from} $1$ {\rm to} $M$}{
            $u \gets$ sample a node in $B$ from distribution $\text{Pr}\{u=u_0\}=\mathbf{d_{u_0}}/s$\;
            $v \gets$ the end of a random walk on $E$ (i.e. graph $G$), starting from $u$ with length $k$\;
            $\widetilde{E} \gets \widetilde{E}\cup \left((u, v), \mathbf{w}_kd_u^{-1/2}d_v^{-1/2}\cdot \frac{s}{M}\right)$ \;
        }
    }
    \Return $\widetilde{E}$\;
\end{algorithm}

\subsection{Node-wise Laplacian Sampling}
\label{app:semi}
In this section, Algorithm~\ref{alg:semi} presents the pseudo-code of the node-wise Laplacian sampling algorithm for semi-supervised tasks.

\begin{table*}[h]
\caption{Detailed information about used datasets.}
\label{tbl:dataset}
\centering
\begin{tabular}{lrrrrrc}
\toprule
Dataset & Nodes & Edges & Features & Classes & $\mathcal{H(G)}$ & scale\\
\midrule
Cora & 2,708 & 5,278 & 1,433 & 7 & 0.810 & small \\
Citeseer & 3,327 & 4,552 & 3,703 & 6 & 0.736 & small \\
PubMed & 19,717 & 44,324 & 500 & 3 & 0.802 & small \\
Actor & 7,600 & 30,019 & 932 & 5 & 0.219 & small \\
Wisconsin & 251 & 515 & 1,703 & 5 & 0.196 & small \\
Cornell & 183 & 298 & 1,703 & 5 & 0.305 & small \\
Texas & 183 & 325 & 1,703 & 5 & 0.108 & small \\
Photo & 7,650 & 119,081 & 745 & 8 & 0.827 & small \\
Computers & 13,752 & 245,861 & 767 & 10 & 0.777 & small \\
Twitch-de & 9,498 & 153,138 & 2,514 & 2 & 0.632 & small \\
Ogbn-arxiv & 169,343 & 1,166,243 & 128 & 40 & 0.655 &medium \\
Twitch-gamers & 168,114 & 6,797,557 & 7 & 2 & 0.554 & medium \\
Penn94 & 41,554 & 1,362,229 & 4,814 & 2 & 0.489 & medium \\
Ogbn-papers100M & 111,059,956 & 1,615,685,872 & 128 & 172 & - & large \\
MAG-Scholar-C & 10,541,560 & 132,609,997 & 2,784,240 & 8 & - & large \\
\bottomrule
\end{tabular}
\end{table*}

\section{Dataset Details}
\label{app:dataset}
All the baselines and our proposed method are tested with various datasets with diverse homophily and scales.
Cora, Citeseer, and PubMed~\citep{DBLP:cora1, DBLP:cora2} have been the most widely tested citation networks since the emergence of the GCN.
Photos and computers~\citep{DBLP:photo1, DBLP:photo2} are the segment of the co-purchase graphs from Amazon.
Actor~\citep{DBLP:Geom-GCN} is the co-occurrence graph of film actors on Wikipedia.
Cornell, Texas, and Wisconsin~\citep{DBLP:Geom-GCN} are the webpage hyperlink graphs from WebKB.
Twitch-de and Twitch-gamers~\citep{DBLP:LINKX} are the social network of Twitch users.
Penn94~\citep{DBLP:LINKX} is a friendship network from the Facebook 100 networks.
Ogbn-arxiv and Ogbn-papers100M~\citep{DBLP:OGB} are two public datasets proposed by the OGB team, where Ogbn-papers100M contains over 100 million nodes and 1 billion edges.
MAG-scholar-C is another benchmark from PPRGo based on Microsoft Academic Graph~\citep{DBLP:MAG} with over 12.4 million nodes and 2.8 million node features.

Here we list the detailed information of used datasets in the experiment.
All the information we proposed here refers to the original version of data collected by PyG.
All the graphs are converted to the undirected graph and added self-loops, which are precomputed and saved.

\section{Additional Experimental results}
\label{app:expres}
\subsection{Applicability and Performance on Multi-layer Models}

Our theoretical analysis is primarily concerned with bounding the error in approximating the polynomial of the Laplacian matrix and the subsequent node representations following the linear transformation stages. 
Importantly, these results are robust and maintain their validity across various model structures and irrespective of the number of layers involved. 
This independence ensures that our theoretical conclusions remain applicable when implementing Laplacian sparsification in any polynomial-based spectral methods. 
Furthermore, although stacking spectral convolution layers is generally not recommended (for extra computational overhead and unstably increasing performance), our findings are persuasive and retain their relevance in such configurations.

In our study, we present the comparative results of multiple configurations of GPR-GNN, including standard GPR-GNN, GPR-GNN enhanced with two layers, GPR-GNN incorporating Laplacian sparsification, and GPR-GNN that combines both multilayer enhancements and sparsification in Table~\ref{tbl:gpr-multi}. 
As our results show, the 2-layer GPR-GNN entangling with Laplacian sparsification also consistently yields improved performance compared to the simply stacked 2-layer GPR-GNN.
Moreover, some result entries of 2-layer GPR-GNN with Laplacian sparsification surpass all the other compared models,
demonstrating the superiority of our methods.
\begin{table*}[htbp]
\caption{The results of GPR-GNN, and its multi-layer / Laplacian sparsification variants. The models with suffix ``-2L'' represent the models are stacked for 2 layers, and those with suffix ``-LS'' represent the models with Laplacian sparsification.}
\label{tbl:gpr-multi}
\centering
\begin{tabular}{lcccccccccc}
\toprule
Dataset & Cora & Citeseer & PubMed & Actor & Wisconsin & Cornell & Texas & Photo & Computers  & Twitch-de\\ 
$\mathcal{H}(G)$ & 0.810 & 0.736 & 0.802 & 0.219 & 0.196 & 0.305 & 0.108 & 0.827 & 0.777 & 0.632 \\
\midrule
GPR-GNN                 & 88.80 & \underline{81.57} & \underline{90.98} & 40.55 & 91.88 & \underline{89.84} & \textbf{92.78} & \underline{95.10} & 89.69  & \textbf{73.90}\\
GPR-GNN-LS              & \textbf{89.31} & \textbf{81.65} & 90.95 & 41.82 & \underline{93.63} & \textbf{91.14} & \underline{92.62} & \textbf{95.30} & \underline{90.47} & 73.71\\
GPR-GNN-2L         & 88.09 & 80.23 & \textbf{91.31} & \underline{42.08} & 92.38 & 85.90 & 87.87 & 94.24 & 90.24 & 73.49\\
GPR-GNN-2L-LS      & \underline{88.82} & 80.60 & \underline{90.98} & \textbf{42.82} & \textbf{96.75} & 89.51 & 91.80 & \underline{95.10} & \textbf{90.57} & \underline{73.88}\\
\bottomrule
\end{tabular}
\end{table*}

\subsection{Comparison among Detached Models}
\label{app:detach}
In this section, we restate some of the results in Table~\ref{tbl:res-small} and provide more experimental tests among GPR-GNN and its variants to argue the potential performance reductions of the detaching trick.
The results are proposed in Table~\ref{tbl:det1} and Table~\ref{tbl:det2}

\begin{table*}[htbp]
\caption{Part of the experimental results in Table~\ref{tbl:res-small}, including the comparison between GCN and SGC.}
\label{tbl:det1}
\centering
\begin{tabular}{lcccccccccc}
\toprule
 Dataset & Cora & Citeseer & PubMed & Actor & Wisconsin & Cornell & Texas & Photo & Computers & Twitch-de \\
$\mathcal{H}(G)$ & 0.810 & 0.736 & 0.802 & 0.219 & 0.196 & 0.305 & 0.108 & 0.827 & 0.777 & 0.632 \\
\midrule
GCN         & 87.78 & 81.50 & 87.39 & 35.62 & 65.75 & 71.96 & 77.38 & 93.62 & 88.98 & 73.72 \\
SGC         & 87.24 & 81.53 & 87.17 & 34.40 & 67.38 & 70.82 & 79.84 & 93.41 & 88.61 & 73.70 \\
$\Delta$    & -0.54 & +0.03 & -0.22 & -1.22 & +1.63 & -1.14 & +2.46 & -0.21 & -0.37 & -0.02 \\
\bottomrule
\end{tabular}
\end{table*}

\begin{table*}[htbp]
\caption{The results of GPR-GNN, GPR-GNN with Laplacian sparsification and detached GPR-GNN.}
\label{tbl:det2}
\centering
\begin{tabular}{lcccccccccc}
\toprule
 Dataset & Cora & Citeseer & PubMed & Actor & Wisconsin & Cornell & Texas & Photo & Computers & twitch-de\\
$\mathcal{H}(G)$ & 0.810 & 0.736 & 0.802 & 0.219 & 0.196 & 0.305 & 0.108 & 0.827 & 0.777 & 0.632 \\
\midrule
GPR-GNN-LS          & \textbf{89.31} & \textbf{81.65} & 90.95 & \textbf{41.82} & \textbf{93.63} & \textbf{91.14} & \underline{92.62} & \textbf{95.30} & \textbf{90.47} & \underline{73.71}\\
\midrule
GPR-GNN             & \underline{88.80} & \underline{81.57} & \textbf{90.98} & \underline{40.55} & \underline{91.88} & \underline{89.84} & \textbf{92.78} & 95.10 & \underline{89.69} & \textbf{73.90} \\
GPR-GNN-detached    & 88.13 & 80.96 & \underline{90.96} & 40.10 & 88.88 & 89.83 & 85.08 & \underline{95.12} & 89.36 & 73.17\\
$\Delta$            & -0.67 & -0.61 & -0.02 & -0.45 & -3.00 & -0.01 & -7.70 & +0.02 & -0.33 & -0.73 \\
\bottomrule
\end{tabular}
\end{table*}

We first examine the comparison of GCN and SGC.
SGC can be considered as the detached version of GCN, which first executes a $K$-hop graph propagation ($K=2$ in practice), followed by the linear layers.
Our empirical findings verify that the detaching manner does exert a negative influence on the performance of GCN.
Despite SGC exhibiting occasional performance improvements on specific small-scale datasets, the overall results are still distant from the powerful baseline GPR.

The variant known as Detached GPR-GNN modifies the original by omitting the linear transformation during graph propagation. 
Although GPR-GNN-detach retains many of GPR-GNN's essential characteristics—thereby significantly outperforming traditional GNNs—it also simplifies the original architecture. 
This simplification could lead to potential performance reductions compared to GPR-GNN, as our proposed results show.

\begin{table*}[htbp]
\caption{The results of GPR-GNN, GPR-GNN with Laplacian sparsification, ClusterGCN, and H2GCN.}
\label{tbl:res-add}
\centering
\begin{tabular}{lcccccccccc}
\toprule
Dataset & Cora & Citeseer & PubMed & Actor & Wisconsin & Cornell & Texas & Photo & Computers  & Penn94\\ 
$\mathcal{H}(G)$ & 0.810 & 0.736 & 0.802 & 0.219 & 0.196 & 0.305 & 0.108 & 0.827 & 0.777 & 0.470 \\
\midrule
GPR-GNN                 & 88.80 & \underline{81.57} & \textbf{90.98} & \underline{40.55} & \underline{91.88} & \underline{89.84} & \textbf{92.78} & \underline{89.69} & \underline{95.10} & \underline{82.92}\\
GPR-GNN-LS              & \textbf{89.31} & \textbf{81.65} & \underline{90.95} & \textbf{41.82} & \textbf{93.63} & \textbf{91.14} & \underline{92.62} & \textbf{90.47} & \textbf{95.30} & \textbf{85.00} \\
ClusterGCN         & 87.45 & 79.66 & 86.52 & 29.66 & 61.88 & 56.72 & 65.08 & 87.11 & 93.17 & 81.75 \\
H2GCN              & -     & -     & 87.78 & 34.49 & 87.50 & 86.23 & 85.90 & -     & -     & 81.31 \\
\bottomrule
\end{tabular}
\end{table*}

\subsection{Additional Comparisons with More Models.}
In this section, we include two more baselines named ClusterGCN~\cite{DBLP:ClusterGCN} and H2GCN~\cite{DBLP:H2GCN}, which are known for their superiority in scalability and unique design for heterophilous graphs.
The results are shown in Table~\ref{tbl:res-add}.
Since we aim to demonstrate that our approach either matches or exceeds the performance of standard spectral GNNS, these models are not aligned with the track of our proposed ones.
We can still show outstanding performance over those scalable methods and spatial ones.

\begin{table}[htbp]
\caption{The time and GPU memory consumption results of APPNP and APPNP-LS on several datasets.}
\label{tbl:timespace}
\centering
\resizebox{\linewidth}{!}{
\begin{tabular}{lcccc}
\toprule
 Time per epoch (ms) & Cora & Computers & Twitch-gamer & Penn94  \\
\midrule
APPNP       & 6.50  & 7.59  & 19.99 & 13.75 \\
APPNP-LS    & 5.58  & 5.83  & 16.73 & 12.42 \\
\midrule
GPU memory (GB) & Cora & Computers & Twitch-gamer & Penn94  \\
APPNP       & 0.034 & 0.144 & 0.976 & 1.725 \\
APPNP-LS    & 0.034 & 0.127 & 0.845 & 1.821 \\
\bottomrule
\end{tabular}
}
\end{table}

\begin{table*}[htbp]
\caption{The executability of different models on the Ogbn-papers100M and MAG-scholar-C.}
\label{tbl:executability}
\centering
\resizebox{\linewidth}{!}{
\begin{tabular}{p{4.3cm}p{2.5cm}p{4.5cm}p{2.5cm}p{4.5cm}}
\toprule
 & Ogbn-papers100M & Reasons for inability to execute & MAG-scholar-C & Reasons for inability to execute  \\
\midrule
GCN, GCNII, GAT\newline (traditional GNNs) & No & GPU OOM during full-batch message-passing & No & GPU OOM during full-batch message-passing\\
\midrule
SGC with precomputed\newline graph propagation& Yes & / & No & OOM during full-batch message-passing on CPU \\
\midrule
APPNP, GPRGNN, BernNet, ChebNetII, JacobiConv, FavardGNN (polynomial-based spectral GNNs) & No & GPU OOM during full-batch message-passing & No & GPU OOM during full-batch message-passing\\
\midrule
Detached spectral GNNs with precomputed graph propagation & Yes & / & No & OOM during full-batch message-passing on CPU \\
\midrule
LazyGNN & No & Error while partitioning the graph & Partial & Executable under extreme parameter settings. It takes a few days to yield unreliable results.\\
\midrule
LMCGCN & No & Error while partitioning the graph & Partial & Executable under extreme parameter settings. It takes extra time and memory to maintain historical embeddings.\\
\midrule
PPRGo & No & Error while calculating the approximation matrix & Yes & / \\
\midrule
Spectral GNNs with our proposed Laplacian sparsification method. & Yes & / & Yes & / \\

\bottomrule
\end{tabular}
}
\end{table*}

\begin{table*}[htbp]
\caption{Experimental results of investigating the impact of hyper-parameter ``--ec''.}
\label{tbl:ec}
\centering
\begin{tabular}{lccccccccccccc}
\toprule
--ec & 0.01 & 0.05 & 0.1 & 0.5 & 1 & 2 & 3 & 4 & 5 & 6 & 7 & 8 & 10 \\
\midrule
PubMed       & 87.30 & 88.96 & 89.50 & 90.49 & 90.80 & 90.95 & 90.98 & 90.84 & 90.88 & \textbf{91.01} & 90.99 & 90.94 & 90.95 \\
Twitch-de    & 70.46 & 71.20 & 72.50 & 72.85 & \textbf{73.40} & 73.19 & 73.15 & 73.31 & 73.17 & 73.12 & 73.19 & 73.22 & 73.26 \\
\bottomrule
\end{tabular}
\end{table*}

\subsection{Analysis on Time, Space, Executability, and Sampling Numbers}
\label{app:scale}
\header{\bf Time, space, and executability.} First, we present some results of our time and memory efficiency comparisons on relatively smaller datasets in Table~\ref{tbl:timespace}.  
These findings will illustrate the enhancements our model offers, making SGNN training viable in wide scenarios.

For runtime and memory consumption, we provide some metrics from tests conducted on the datasets MAG-scholar-C and Ogbn-papers100M. 
Compared to traditional SGNNs, our model significantly reduces memory overhead to $O(Ln_bF + LF^2 + F_iF + m)$ per batch, where $n_b$ is the number of nodes per batch, and $F_i$ denotes the dimensions of the original node embeddings.
This reduction enables the practical training of SGNNs on the Ogbn-papers100M dataset using Laplacian sparsification.

Additionally, unlike detached SGNNs, which face infeasibility issues due to the high memory demands of propagating node embeddings on CPU ($O(KnF_i)$ on the MAG-scholar-C dataset), our method reduces this overhead and makes training feasible on such large-scale datasets. 
This transition from impractical to applicable training scenarios underlines our model's improvements in memory management and computational efficiency.
For more information about time and memory, we recommend the readers refer to the time and space complexity analysis in Appendix~\ref{app:overheadanalysis} for a comprehensive understanding.

To demonstrate the universality of our methods in large-scale tasks, we present Table~\ref{tbl:executability} to show the excitability of different models on the Ogbn-papers100M and MAG-scholar-C datasets, along with the reasons for any model's inability to execute.

% To the best of our knowledge, as PPRGo suffers runtime error when dealing with dataset Ogbn-papers100M, our model is one of the few that can simultaneously overcome the scalability obstacle raised by dataset Ogbn-papers100M and MAG-scholar-C.

\header{\bf Sampling Numbers.} The relationship between approximation similarity—controlled by sampling numbers—and final GNN performance is complex. 
While insufficient sampling numbers can disrupt the graph filter significantly, leading to unacceptable inaccuracies, increasing the number of samples does not invariably enhance performance. 
In some cases, an optimal level of randomness can actually mitigate the effects of noise inherent in real-world datasets. This interaction is evidenced by the variability in optimal settings for the ``--ec'' parameter. 
Notably, the best parameters do not uniformly include ``--ec=10'', indicating that more sampling is not always beneficial.

We present some results from our ablation study in Table~\ref{tbl:ec}, which investigates the relationship between approximation similarity and the ultimate performance of GNN models. 
We have selected the GPR-GNN-LS model as our primary focus for this study. 
As the results show, when ``--ec'' is at a low level (<1), the performance increases significantly as ``--ec'' increases.
The performance then stabilizes without severe fluctuations, no matter how the growth of ``--ec'' strengthens the similarity of approximation.
The findings aim to clarify how different settings of the ``--ec'' parameter influence model effectiveness, providing insights that could guide the optimization of sampling strategies in practical GNN applications.
In reality, we may limit the ``--ec'' to a reasonable range, such as $[1, 10]$, and perform hyper-parameter tuning, similar to how the ``learning rate'' is selected.

\section{Experiment Details}
\label{app:params}
\subsection{Codes}
Our codes are released at \url{https://anonymous.4open.science/r/SGNN-LS-release-B926}.

\subsection{Configurations}
\label{app:config}
Here we list the detailed information on the experimental platform and the environment we deployed.
\begin{itemize}
    \item Operating System: Red Hat Linux Server release 7.9 (Maipo).
    \item CPU: Intel(R) Xeon(R) Gold 8358 64C@2.6GHz.
    \item GPU: NVIDIA A100 40GB PCIe. \\ GPU: NVIDIA A100 80GB PCIe for the experiment on Ogbn-papers100M.
    \item GPU Driver: 525.125.06.
    \item RAM: 512G.
    \item Python: 3.10.6.
    \item CUDA toolkit: 11.3.
    \item Pytorch: 1.12.1.
    \item Pytorch-geometric: 2.1.0.
\end{itemize}

\begin{table*}[h]
\caption{The hyper-parameters of APPNP-LS on small-scale datasets.}
\label{tbl:appnp-hyp-small}
\centering
\begin{tabular}{lcccccc}
\toprule
Dataset & learning rate & weight decay & dropout=dprate & $\alpha$ & $K$ & ec \\ 
\midrule
Cora & 0.05 & 0.0005 & 0.5 & 0.1 & 5 & 10 \\
Citeseer & 0.01 & 0 & 0.8 & 0.5 & 10 & 10 \\
PubMed & 0.05 & 0.0005 & 0.0 & 0.5 & 10 & 10\\
Actor & 0.01 & 0.0005 & 0.8 & 0.9 & 2 & 10\\
Wisconsin & 0.05 & 0.0005 & 0.5 & 0.9 & 10 & 20\\
Cornell & 0.05 & 0.0005 & 0.5 & 0.9 & 5 & 20\\
Texas & 0.05 & 0.0005 & 0.8 & 0.9 & 2 & 20\\
Photo & 0.05 & 0 & 0.5 & 0.5 & 5 & 10\\
Computers & 0.05 & 0 & 0.2 & 0.1 & 2 & 10\\
Twitch-de & 0.01 & 0.0005 & 0.5 & 0.1 & 2 & 10\\
\bottomrule
\end{tabular}
\end{table*}
\begin{table*}[h]
\caption{The hyper-parameters of GPR-LS on small-scale datasets.}
\label{tbl:gpr-hyp-small}
\centering
\begin{tabular}{lcccccc}
\toprule
Dataset & lr=prop\_lr & wd=prop\_wd & dropout=dprate & $\alpha$ & $K$ & ec \\ 
\midrule
Cora & 0.05 & 0.0005 & 0.8 & 0.9 & 10 & 3 \\
Citeseer & 0.01 & 0 & 0.2 & 0.5 & 10 & 10 \\
PubMed & 0.05 & 0.0005 & 0.2 & 0.5 & 5 & 10\\
Actor & 0.05 & 0.0005 & 0.5 & 0.5 & 10 & 3\\
Wisconsin & 0.05 & 0.0005 & 0.8 & 0.9 & 10 & 10\\
Cornell & 0.05 & 0.0005 & 0.5 & 0.9 & 2 & 1\\
Texas & 0.05 & 0.0005 & 0.8 & 0.5 & 10 & 10\\
Photo & 0.05 & 0.0005 & 0.8 & 0.1 & 2 & 10\\
Computers & 0.05 & 0 & 0.5 & 0.1 & 10 & 10\\
Twitch-de & 0.01 & 0.0005 & 0.8 & 0.1 & 2 & 20\\
\bottomrule
\end{tabular}
\end{table*}
\begin{table*}[h]
\caption{The hyper-parameters of Jacobi-LS on small-scale datasets.}
\label{tbl:jacobi-hyp-small}
\centering
\begin{tabular}{lccccccc}
\toprule
Dataset & lr=prop\_lr & wd=prop\_wd & dropout=dprate & paraA & paraB & $K$ & ec \\ 
\midrule
Cora & 0.05 & 0.0005 & 0.8 & 0.5 & 0.5 & 10 & 3 \\
Citeseer & 0.01 & 0 & 0 & 0.5 & 0.5 & 2 & 1 \\
PubMed & 0.05 & 0.0005 & 0.2 & 1.0 & 0.5 & 5 & 3\\
Actor & 0.01 & 0.0005 & 0.5 & 0.5 & 0.5 & 10 & 1\\
Wisconsin & 0.05 & 0.0005 & 0.8 & 0.5 & 0.5 & 2 & 10\\
Cornell & 0.05 & 0.0005 & 0.8 & 0.5 & 0.5 & 2 & 1\\
Texas & 0.05 & 0.0005 & 0.5 & 0.5 & 1.0 & 10 & 1\\
Photo & 0.05 & 0 & 0.5 & 1.0 & 0.5 & 10 & 3\\
Computers & 0.05 & 0.0005 & 0.2 & 1.0 & 1.0 & 2 & 3\\
Twitch-de & 0.01 & 0.0005 & 0.8 & 0.5 & 1.0 & 5 & 10\\
\bottomrule
\end{tabular}
\end{table*}
\begin{table*}[h]
\caption{The hyper-parameters of Favard-LS on small-scale datasets.}
\label{tbl:favard-hyp-small}
\centering
\begin{tabular}{lccccc}
\toprule
Dataset & lr=prop\_lr & wd=prop\_wd & dropout=dprate & $K$ & ec \\ 
\midrule
Cora & 0.05 & 0.0005 & 0.8 & 2 & 1 \\
Citeseer & 0.01 & 0.0005 & 0 & 2 & 1 \\
PubMed & 0.01 & 0.0005 & 0.2 & 5 & 3\\
Actor & 0.05 & 0.0005 & 0 & 2 & 1\\
Wisconsin & 0.05 & 0.0005 & 0.8 & 5 & 20\\
Cornell & 0.05 & 0.0005 & 0.2 & 5 & 1\\
Texas & 0.05 & 0.0005 & 0.8 & 10 & 10\\
Photo & 0.05 & 0.0005 & 0.8 & 2 & 10\\
Computers & 0.05 & 0.0005 & 0.8 & 5 & 10\\
Twitch-de & 0.01 & 0 & 0.8 & 10 & 10\\
\bottomrule
\end{tabular}
\end{table*}
\begin{table*}[h]
\caption{The hyper-parameters of APPNP-LS on medium- and large-scale datasets.}
\label{tbl:appnp-hyp-big}
\centering
\begin{tabular}{lcccccc}
\toprule
Dataset & learning rate & weight decay & dropout=dprate & $\alpha$ & $K$ & ec \\ 
\midrule
Ogbn-arxiv & 0.01 & 0 & 0 & 0.1 & 5 & 10 \\
Twitch-gamer & 0.01 & 0 & 0 & 0.5 & 2 & 10 \\
Penn94 & 0.01 & 0 & 0.5 & 0.9 & 10 & 10 \\
Ogbn-papers100M & 0.01 & 0 & 0.2 & 0.1 & 2 & - \\
MAG-scholar-C & 0.01 & 0 & 0.5 & 0.5 & 2 & - \\
\bottomrule
\end{tabular}
\end{table*}
\begin{table*}[h]
\caption{The hyper-parameters of GPR-LS on medium- and large-scale datasets.}
\label{tbl:gpr-hyp-big}
\centering
\begin{tabular}{lcccccc}
\toprule
Dataset & lr=prop\_lr & wd=prop\_wd & dropout=dprate & $\alpha$ & $K$ & ec \\ 
\midrule
Ogbn-arxiv & 0.01 & 0 & 0.2 & 0.5 & 10 & 10 \\
Twitch-gamer & 0.05 & 0 & 0.5 & 0.1 & 5 & 10 \\
Penn94 & 0.01 & 0 & 0.5 & 0.1 & 2 & 10 \\
Ogbn-papers100M & 0.01 & 0 & 0.2 & 0.9 & 2 & - \\
MAG-scholar-C & 0.01 & 0 & 0.5 & 0.5 & 2 & -\\
\bottomrule
\end{tabular}
\end{table*}

\subsection{Details for small-scale experiments.}
For all the datasets involved in Table~\ref{tbl:res-small}, we conduct a full-supervised node classification experiment.
Except for Twitch-de, all the datasets are randomly divided into 10 splits with the commonly adapted training/validation/test proportions of 60\%/20\%/20\%.
For the Twitch-de, we use the default 5 splits proposed by PyG and propose the average performance of all the datasets.

We limit the hidden size to 64 and the layers of MLP to 2 for a relatively fair comparison among all the tested models. 
We propose the best hyperparameters of our Laplacian sparsification extended models in Table~\ref{tbl:appnp-hyp-small},~\ref{tbl:gpr-hyp-small},~\ref{tbl:jacobi-hyp-small}, and~\ref{tbl:favard-hyp-small} for full reproducibility.

\subsection{Details for large-scale experiments}
For all the datasets involved in Table~\ref{tbl:res-large} except MAG-scholar-C, we conduct a supervised node classification experiment.
Since the proportion of training nodes in Ogbn-papers100M is small, we consider it a semi-supervised task.
All the datasets here have the default splits provided by PyG.
For Twitch-gamers and Penn94, we propose the average accuracy of all the default splits.
For OGB datasets, we repeat the experiment on the single split 5 times with different random seeds and report the average accuracy.

For MAG-scholar-C, we continue to use the academic settings provided in PPRGo, which sets the size of the training/validation sets to 105415 nodes.
We repeat the experiment on the randomly generated splits with 5 fixed random seeds and report the average accuracy.

We limit the hidden size to 128 and the layers of MLP to 3 at most for a relatively fair comparison among all the tested models.
We propose the best hyperparameters for our Laplacian sparsification extended models in Table ~\ref{tbl:appnp-hyp-big} and ~\ref{tbl:gpr-hyp-big} for full reproducibility.

\end{document}